\documentclass{article}
\usepackage{ranger_preprint,times}

\usepackage{amsmath,amsfonts,bm}

\def\eqref#1{equation~\ref{#1}}

\def\1{\bm{1}}

\DeclareMathAlphabet{\mathsfit}{\encodingdefault}{\sfdefault}{m}{sl}
\SetMathAlphabet{\mathsfit}{bold}{\encodingdefault}{\sfdefault}{bx}{n}

\usepackage{hyperref}
\usepackage{url}
\hypersetup{
  pdftitle={Agentic Image Generation in the Wild},
  pdfauthor={Yifan Li, Bowen Qin, Kun Zhou, Haochen Zhao, Zikang Liu, Zheng Li, Minsheng Hao, Yuhao Cheng, Haosheng Zou, Jeff Chen, Wayne Xin Zhao}
}
\usepackage{graphicx}
\usepackage{float}
\usepackage{wrapfig}
\usepackage{booktabs}
\usepackage{multirow}
\usepackage{amsmath}
\usepackage{amssymb}
\usepackage{xcolor}
\usepackage{colortbl}
\usepackage{subcaption}
\usepackage[capitalize]{cleveref}
\usepackage{xspace}
\usepackage{longtable}
\usepackage{listings}
\lstdefinestyle{judgejson}{
  basicstyle=\footnotesize\ttfamily,
  columns=fullflexible,
  keepspaces=true,
  breaklines=true,
  breakatwhitespace=true,
  showstringspaces=false,
  aboveskip=4pt,
  belowskip=2pt,
  frame=single,
  rulecolor=\color{black!18},
  backgroundcolor=\color{black!3},
  framerule=0.3pt,
  framesep=5pt,
  xleftmargin=6pt,
  xrightmargin=6pt
}

\graphicspath{{figures/}}

\newcommand{\savanna}{Savanna}
\newcommand{\ranger}{Ranger}
\newcommand{\ourbench}{\savanna}
\newcommand{\ourdata}{\mbox{Ranger-15K}}
\newcommand{\ourmodel}{\ranger}

\newcommand{\eg}{\emph{e.g.,}\xspace}

\title{Agentic Image Generation in the Wild}

\author{%
\textbf{Yifan Li}$^{1,*}$, \textbf{Bowen Qin}$^{2}$, \textbf{Kun Zhou}$^{1}$, \textbf{Haochen Zhao}$^{3}$, \textbf{Zikang Liu}$^{1}$, \textbf{Zheng Li}$^{3}$, \\
\textbf{Minsheng Hao}$^{2}$, \textbf{Yuhao Cheng}$^{2}$, \textbf{Haosheng Zou}$^{2}$, \textbf{Jeff Chen}$^{2,\ddagger}$, \textbf{Wayne Xin Zhao}$^{1,\dagger}$  \\
$^{1}$Renmin University of China \quad $^{2}$Tencent \quad $^{3}$Peking University \\
\texttt{\{liyifan0925\}@gmail.com}
}

\begin{document}
\raggedbottom

\maketitle
\lhead{Preprint.}
{\renewcommand{\thefootnote}{\fnsymbol{footnote}}%
\footnotetext[1]{Work done during internship at Tencent.}%
\footnotetext[2]{Corresponding author.}%
\footnotetext[3]{Project leader.}}

\begin{abstract}

Agentic image generation pairs an image generator with a search agent that retrieves references from the web, extending generation beyond the generator's training data. Real-world requests, however, often concern entities that appeared only days ago. We call this setting Image Generation in the Wild. In this setting, the web offers limited visual evidence, search results are crowded with easily confused entities, and the target is rarely shown on its own. To study this setting, we build an automated pipeline that synthesizes prompts from recent headlines and keeps only those whose real search results are highly confusing, yielding requests that are hard to solve yet natural to ask. With it, we construct \ourbench{}, a benchmark of 500 prompts across 16 subcategories. On \ourbench{}, even harness frameworks built on the frontier models retrieve a correct reference for only about half of the prompts. We then propose Locate-Corroborate-Isolate, a paradigm in which the agent first pins down the target's identity through text search, then verifies candidate images against multiple sources, and finally crops a clean reference for the generator. Trained on 15K trajectories that follow this paradigm, \ourmodel{}-8B achieves a search score of 67.2, outperforming harnesses equipped with frontier models. It also achieves the best generation quality with every generator we test, and surpasses prior agentic methods on other image generation benchmarks. We release the pipeline, benchmark, training data and model at \url{https://github.com/RUCAIBox/Agentic-ImageGen-Wild}.

\end{abstract}

\section{Introduction}
\label{sec:intro}
\begin{figure}[t]
    \centering
    \includegraphics[width=\linewidth]{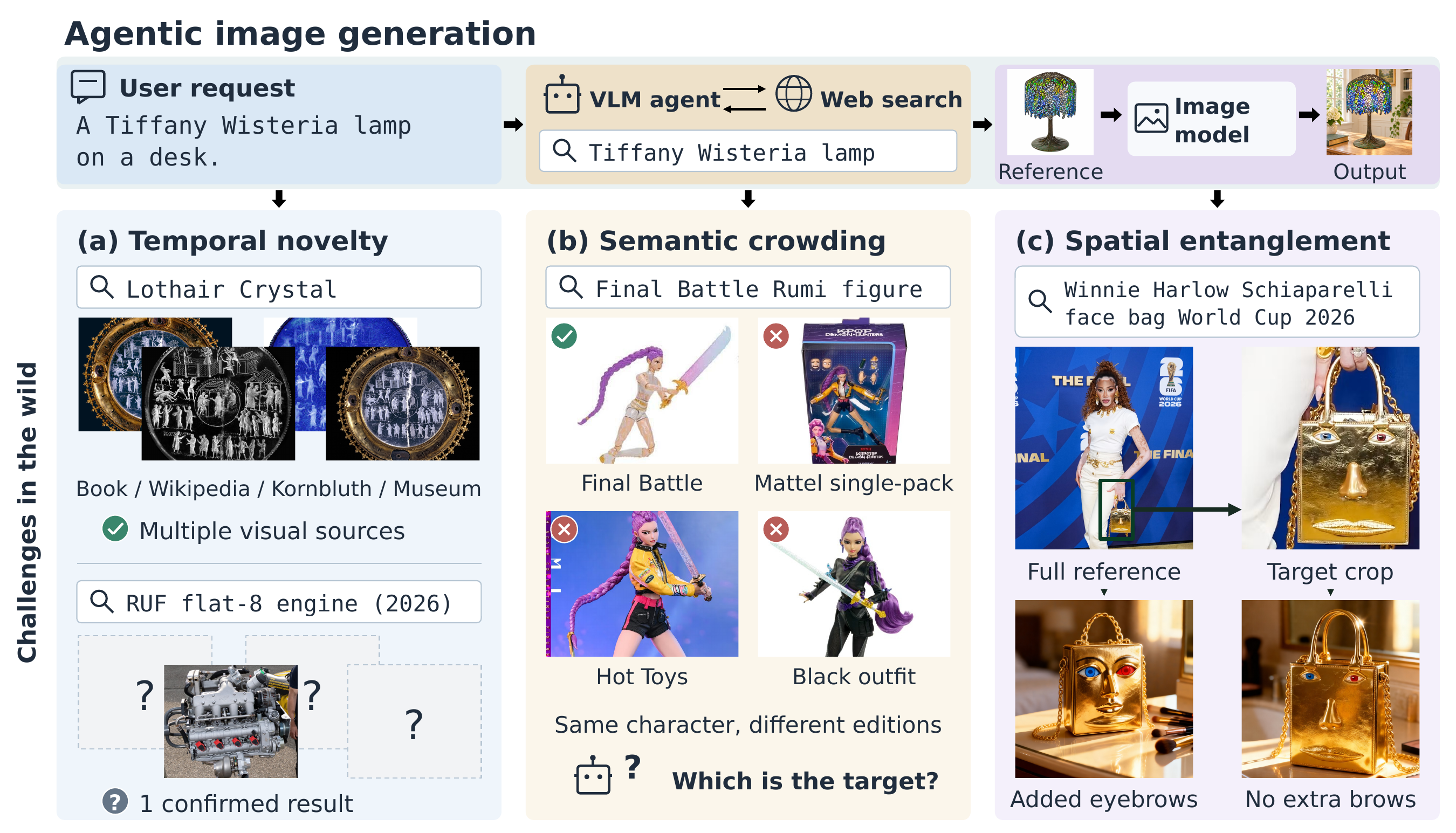}
    \caption{\textbf{Agentic image generation and its challenges in the wild.}
    Top: the agentic image generation workflow.
    Bottom: (a) temporal novelty, (b) semantic crowding, and (c) spatial
    entanglement.}
    \label{fig:in_the_wild}
\end{figure}

Generative models~\citep{wu2025qwenimage,google2025nanobananapro} have made substantial progress in generating high-quality images from natural-language prompts. Agentic image generation~\citep{feng2026gensearcher,chen2026genevolve} further extends these capabilities by equipping vision-language model (VLM) agents with web search tools to retrieve textual and visual references, improving performance on tasks that require reasoning or knowledge of uncommon entities~\citep{niu2026wise,he2026mindbrush}. Despite this progress, current systems still struggle in realistic settings, which we refer to as \textbf{Image Generation in the Wild}. We identify three major challenges in this setting (\cref{fig:in_the_wild}). \textbf{Temporal novelty:} users may request images of entities shortly after their introduction. Unlike an established uncommon entity, a newly introduced one lacks a stable visual record, making its appearance harder to confirm. \textbf{Semantic crowding:} web searches often return entities that are easily confused with the target. For example, a search for a newly released car may also return its earlier versions, leading agents to select incorrect references. \textbf{Spatial entanglement:} available images may show the target only together with other objects or against a distracting background, which interferes with downstream generation.

To study and address these challenges, we introduce an end-to-end framework for agentic image generation in the wild, spanning data synthesis, evaluation, and model training. We first build an automated pipeline that turns recent headlines into image generation prompts that exhibit these challenges. It starts by collecting recent news and related reports into \emph{world snapshots}, which serve as the source of newly introduced entities. To mimic the web that a time-sensitive request faces, all later searches are restricted to results published within three days of each article. The pipeline then extracts target entities from each snapshot, either from the text or from objects that appear only in the retrieved images. Next, it measures the \emph{retrieval confusion} of each entity and keeps only entities whose search results are sufficiently confusing. A two-stage VLM verification then ensures that each remaining entity points to a single, unambiguous answer. Finally, each entity is written into a prompt with a ground-truth reference and evaluation rubrics. Together, these designs reproduce temporal novelty, semantic crowding, and spatial entanglement in the resulting prompts, which still read like natural user requests and can be synthesized continually as long as news keeps arriving.

Using this pipeline, we construct \textbf{\ourbench}, a benchmark of 500 prompts spanning 16 subcategories in five domains. We evaluate retrieval and generation separately: the \emph{search score} measures whether the agent retrieves correct visual references for the target, and the \emph{generation score} measures whether the final image faithfully depicts the target in the requested context. From proprietary generators to agentic post-trained models, all systems struggle on \ourbench{}. Even harness frameworks~\citep{ren2026scope,ye2026genclaw} built on the latest frontier models such as GPT-6-Astra~\citep{openai2026gpt6astra} and Claude-Opus-5~\citep{anthropic2026opus5} reach a search score of at most 51.5. 

These results suggest that simply collecting search results and passing them to the generator is insufficient in the wild. We therefore propose the \emph{Locate-Corroborate-Isolate} paradigm.
First, in the Locate phase, the agent uses text search and web fetching to establish the target's identity before searching for images, compensating for the missing visual record. Next, in the Corroborate phase, the agent carefully checks each candidate image against the previously gathered facts and other candidates, addressing semantic crowding. Finally, in the Isolate phase, the agent crops the image when necessary into a clean reference, addressing spatial entanglement. To teach this paradigm, we reuse our pipeline to synthesize training prompts and use a teacher model to solve them following the paradigm. We finally keep 15K trajectories whose final references match the ground truth. Fine-tuned on these trajectories, \textbf{\ourmodel{}-8B} reaches a search score of 67.2, 15.7 points above the strongest baseline and more than 20 points above agents with the same backbone. It also achieves the highest generation score with every downstream generator we test, and outperforms prior agentic methods on Mind-Bench~\citep{he2026mindbrush} and KnowGen~\citep{feng2026gensearcher}. Further analysis validates the effectiveness of our design in addressing these challenges.

\section{Savanna: Is Image Generation Ready for the Real World?}
\label{sec:benchmark}

This section first formulates agentic image generation and the challenges it faces in the wild (\cref{sec:task_formulation}). We then describe how we synthesize queries that exhibit these challenges (\cref{sec:query_synthesis}) and how we curate and evaluate \ourbench{} (\cref{sec:benchmark_curation}).

\subsection{Task Formulation}
\label{sec:task_formulation}
\paragraph{Agentic image generation.}
An image generator can only depict entities it has learned, yet users often ask for entities it has never seen. Agentic image generation closes this gap by placing an agent $\pi$ before the generator. Given a user prompt $p$ that asks for a newly introduced entity $e$, the agent searches the web $\mathcal{W}$ for textual and visual knowledge about $e$. It then outputs a refined prompt $\hat{p}$ and a set of reference images $\mathcal{I}_{\mathrm{ref}}$, from which a generator $G$ produces the final image $I_{\mathrm{gen}}$:
\begin{equation}
    (\hat{p}, \mathcal{I}_{\mathrm{ref}}) = \pi(p;\, \mathcal{W}), \qquad
    I_{\mathrm{gen}} = G(\hat{p}, \mathcal{I}_{\mathrm{ref}}).
    \label{eq:agentic_generation}
\end{equation}

\paragraph{The Call of the Wild.}
In real-world scenarios, users often ask for entities that have only just appeared, and the agent can only search the web as it is at that moment. Let $\tau$ denote the time a prompt is issued, and $\mathcal{I}_e$ the images on $\mathcal{W}$ that depict $e$. We observe three core challenges in this setting, illustrated in \cref{fig:in_the_wild}.
\begin{itemize}
    \item \textbf{Temporal novelty.} When $e$ is introduced shortly before $\tau$, the web has had little time to catch up. Its images may not yet be indexed by search engines, so a direct image search for $e$ often misses them. Even when some are found, $\mathcal{I}_e$ is too small to form a visual consensus, making it hard to tell which retrieved image actually depicts $e$ (\cref{fig:in_the_wild}a).
    \item \textbf{Semantic crowding.} A newly introduced entity rarely stands alone in search results. Instead, its results are mixed with other entities that share its name or belong to the same product line. For example, a search for a club's new home kit also returns last season's kit, and a search for a new film returns posters of an older film with the same title. Such results easily mislead the agent into selecting the wrong entity (\cref{fig:in_the_wild}b).
    \item \textbf{Spatial entanglement.} Many entities are photographed only together with other objects or against a distracting background, such as one product among several on a launch stage. Even an image that correctly depicts $e$ may then be unsuitable as a reference, since the generator tends to reproduce everything it contains (\cref{fig:in_the_wild}c).
\end{itemize}
Together, these challenges make a correct and clean reference for $e$ hard to obtain.

\subsection{From Headlines to Hard Prompts}
\label{sec:query_synthesis}
\begin{figure}[!t]
    \centering
    \includegraphics[width=\linewidth]{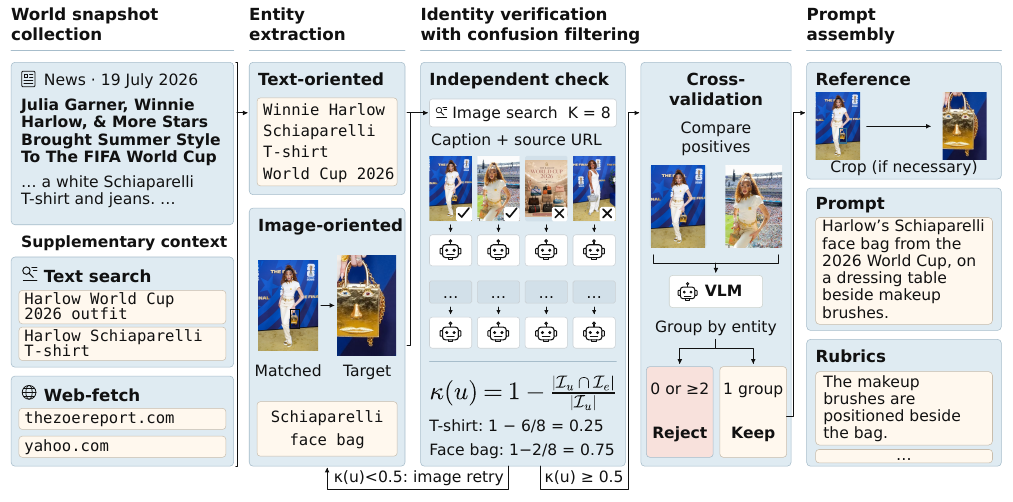}
    \caption{\textbf{Prompt synthesis pipeline}, illustrated with real retrieval results. The text-oriented T-shirt query is
    too easy ($\kappa=0.25$), so image-oriented extraction proposes the face bag
    ($\kappa=0.75$).}
    \label{fig:query_synthesis_pipeline}
\end{figure}

Studying image generation in the wild calls for a steady supply of prompts that exhibit these challenges, both to evaluate agents and to train them. The most direct approach is to collect and filter real user prompts, but this requires access to a large and steady stream of user requests. Can similar prompts instead be constructed automatically from scratch? Our answer is an automated pipeline that turns recent headlines into hard prompts, illustrated in \cref{fig:query_synthesis_pipeline} and described below.

\paragraph{World snapshot collection.}
The very first step is to collect recent news articles. We predefine a set of domain-specific meta queries and use them to search for news day by day over a given time range, recording the title, snippet, source, and publication date of each article. 
Next, we enrich the context of each article by fetching its full text and running another text search on its title to gather related reports from other sources. The article and this supplementary context together form a \emph{world snapshot}: a record of what the web says about an event shortly after it happens. Specifically, we treat the news date as the issue time $\tau$ and restrict all subsequent text or image searches to results published no later than three days after $\tau$, thereby forming a web environment with temporal novelty. Implementation details are presented in \cref{sec:app_tools}.

\paragraph{Entity extraction.}
Given a world snapshot, we can now derive the target entity $e$ that a prompt $p$ will ask for. We extract entities in two ways, called text-oriented and image-oriented, respectively. In \emph{text-oriented} extraction, an LLM directly extracts an entity from the world snapshot (\eg the Schiaparelli T-shirt that Winnie Harlow wore to the 2026 World Cup). We require the entity to be visually identifiable and to carry distinctive attributes, so that it is worth retrieving and can be used to judge whether a generated image is correct. However, an entity reported in the text is usually shown as the main subject in news photos, which naturally provide clean references. To introduce spatial entanglement, we further perform \emph{image-oriented} extraction: a VLM checks the images retrieved for a text-oriented entity and picks another entity that is visible in these images but not mentioned in the text (\eg the face-shaped bag in \cref{fig:query_synthesis_pipeline}). Finally, each extracted entity is written as an image search query $u$ and passed to the next step.

\paragraph{Identity verification with confusion filtering.}
Given the search queries, we next identify which retrieved images truly depict each target entity, filter out queries that are too easy, and ensure that each remaining query has exactly one correct answer. Specifically, for each query $u$ we retrieve a set $\mathcal{I}_u$ of $K=8$ images. These images first go through the \emph{independent check} phase, where a VLM judges each image separately, together with its caption and source URL. We then define the \emph{retrieval confusion} $\kappa(u)$ as the fraction of results that do not depict $e$:
\begin{equation}
    \kappa(u) = 1 - \frac{\left|\mathcal{I}_u\cap\mathcal{I}_e\right|}{\left|\mathcal{I}_u\right|}.
    \label{eq:retrieval_confusion}
\end{equation}
We keep only queries with $\kappa(u)\ge0.5$, so that at least half of the search results show a wrong entity, thereby ensuring semantic crowding around each target entity. Text-oriented queries below the threshold are considered too easy and are sent back to the previous phase for image-oriented extraction. For the remaining queries, the images in $\mathcal{I}_u\cap\mathcal{I}_e$ are sent to the \emph{cross-validation} phase, where a VLM compares them jointly and groups them by whether they show the same entity. We discard queries that match no group, which have no answer, and those that match two or more groups, which are ambiguous. The images in the single remaining group form the candidate set $\mathcal{I}_{\mathrm{cand}}$, from which the reference images are later selected.

\paragraph{Prompt assembly.}
At last, each remaining query is turned into an image generation prompt, paired with a ground-truth reference and rubrics for evaluation. First, a VLM selects the image in $\mathcal{I}_{\mathrm{cand}}$ that shows $e$ most clearly, and crops $e$ from it when necessary, forming the reference images $\mathcal{I}_{\mathrm{ref}}$. Then, a VLM expands $e$ into a prompt $p$ that places $e$ in a new scene and describes its interaction with the surroundings, without revealing its appearance. Finally, the scene information in $p$ is extracted as rubrics for the subsequent evaluation.

Together, these steps produce prompts that concern newly introduced entities, are difficult under real search conditions, and have a single verified answer, while remaining close to what a user would actually ask. The pipeline can also keep producing prompts as long as news keeps arriving.

\begin{figure}[t]
    \centering
    \includegraphics[width=\linewidth]{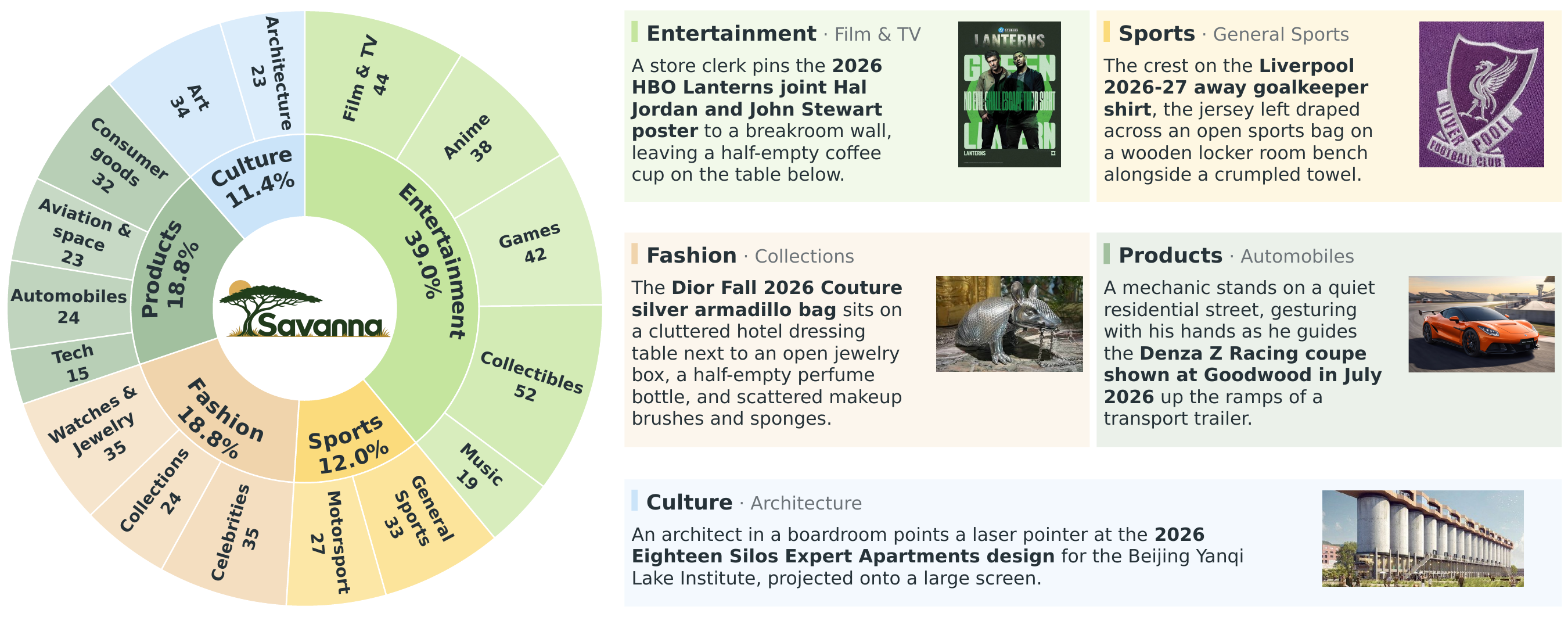}
    \caption{\textbf{Overview of \ourbench{}.} The rings show the five domains (inner) and 16 subcategories (outer). Each domain has one example prompt with its reference image.}
    \label{fig:savanna_benchmark_overview}
\end{figure}
\subsection{Benchmark Curation}
\label{sec:benchmark_curation}
Following this pipeline, we build our benchmark, \textbf{\ourbench{}}. We select five domains that users frequently ask about, namely Entertainment, Sports, Fashion, Products, and Culture, covering 16 subcategories. For each subcategory, we design several meta queries, for a total of 50 (fully presented in \cref{sec:app_meta_query}). We restrict news articles to July 2026, yielding 500 prompts, 399 from text-oriented and 101 from image-oriented extraction. \Cref{fig:savanna_benchmark_overview} shows their distribution over domains and subcategories. Finally, three PhD-level annotators independently review every sample, ensuring that it has exactly one correct answer and the reference correctly corresponds to the target entity.
\paragraph{Metrics.}
We evaluate agents with two scores. The \emph{search score} is the fraction of prompts whose retrieved references match the ground-truth reference of $e$. The \emph{generation score} rates $I_{\mathrm{gen}}$ mainly on whether $e$ is drawn consistently with the ground-truth reference, and partly on whether the context in $p$ satisfies the rubrics. Both scores are evaluated by a VLM. Details are given in \cref{sec:app_metrics}. 

\section{\ourmodel: Leading Image Generation Through the Wild}
\label{sec:method}

Each challenge in the real world calls for a dedicated step in the search. Accordingly, we propose a reasoning paradigm that \emph{locates} the identity of $e$ from text before searching for images, \emph{corroborates} candidate images against their sources and against each other, and \emph{isolates} the verified target into a clean reference (\cref{sec:agent_loop}). Following this paradigm, we train \textbf{Ranger-8B}, a VLM agent learned from teacher trajectories on synthesized prompts (\cref{sec:training}).

\subsection{Locate, Corroborate, Isolate: Surviving the Wild in Three Steps}
\label{sec:agent_loop}

To let the agent gather knowledge from the web, we first build an agent environment with four tools: \texttt{text\_search} and \texttt{web\_fetch} for textual evidence, \texttt{image\_search} for candidate images, and \texttt{crop\_image} for cutting a region out of a retrieved image. On top of this environment, we design a three-step reasoning process. Tool implementations are presented in \Cref{sec:app_tools}.

\begin{figure}[!htbp]
    \centering
    \includegraphics[width=\linewidth]{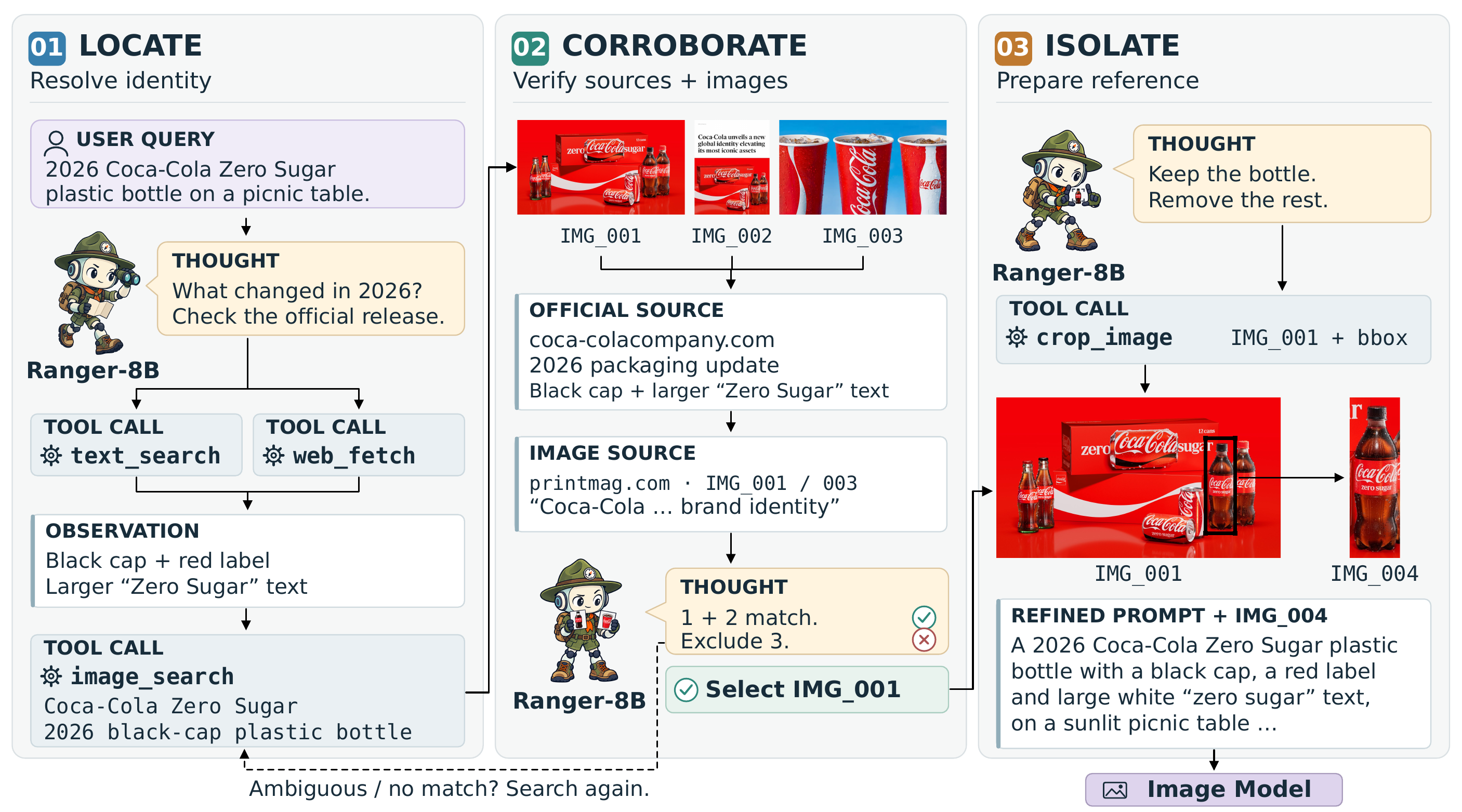}
    \caption{\textbf{The Locate--Corroborate--Isolate paradigm.}
    The agent identifies $e$ from text, verifies candidates against sources and
    each other, then crops a clean reference for the generator.}
    \label{fig:ranger_trajectory}
\end{figure}

\paragraph{Locate.}
A natural first move is to search for images of $e$ directly with the prompt. For an entity that has only just appeared, however, this is unreliable: its images may not yet be indexed, and the few that are found offer no visual consensus for telling which of them shows $e$. The prompt offers little help with either problem, as it names $e$ but says little about what sets it apart. The agent therefore turns to text first, since textual reports can supply exactly these details. It uses \texttt{text\_search}, and \texttt{web\_fetch} when snippets are insufficient, to gather what is known about $e$, such as its maker, release event, and distinguishing features. These facts serve two purposes: they yield an image search query more precise than the prompt, and they provide the criteria for judging candidates in the next step. In \cref{fig:ranger_trajectory}, the agent learns from the official release that the 2026 Zero Sugar bottle has a black cap and an updated label before issuing its first image search.

\paragraph{Corroborate.}
A precise query narrows the search, but does not clean it up. The results remain crowded with entities that share the name or product line of $e$, each of which can look plausible on its own. Rather than trusting any single image, the agent corroborates each candidate in two ways. First, it checks the candidate \emph{across modalities}: besides the pixels, it reads the accompanying text, such as the image title and source page, and compares both with the facts from Locate, rejecting candidates that show a different version, colorway, or member of the same product line. Second, when several candidates remain, it checks them \emph{across images}, looking for a group that shows the same entity. Such agreement, when present, adds confidence that no wrong entity has slipped through. If no candidate remains or several groups are evenly matched, the agent returns to Locate to gather more context about $e$, and then searches again with a more distinctive query.

\paragraph{Isolate.}
A verified image shows the right entity, but not necessarily that entity alone. It may also contain other objects or a distracting background. Given such a reference, the generator may confuse which object the prompt refers to, or carry the extra content into its output, degrading the generated image. The last step therefore disentangles $e$ from its surroundings. The agent selects the verified image whose view best matches $p$, and when $e$ shares the frame with other objects, calls \texttt{crop\_image} with a bounding box around $e$, which returns a new, cropped image with its own ID. In \cref{fig:ranger_trajectory}, the agent crops the bottle out of a verified product lineup that also includes cans and glass bottles, and passes the crop to the generator together with $\hat{p}$.

\subsection{Learning from Teacher Trajectories}
\label{sec:training}
The three stages require behaviors that current open models rarely show. We first collect teacher trajectories and train \ourmodel{}-8B.

\paragraph{Trajectory collection.}
We run the pipeline in \cref{sec:query_synthesis} on news from January to June 2026 to obtain training prompts, which are disjoint in time from \ourbench{}, avoiding data leakage. We then adopt Gemini-3.1-pro~\citep{google2026gemini31pro} as the teacher model to solve each prompt in the same agent environment, with the proposed reasoning paradigm. We keep only trajectories whose final reference matches the ground truth, and discard those whose $\hat{p}$ drops conditions of $p$ or that repeat searches without gaining new information. The remaining 15,119 trajectories form \textbf{\ourdata{}}.

\paragraph{Training details.}
Utilizing these trajectories, we train \ourmodel{}-8B on \ourdata{} using 8 NVIDIA H20 GPUs, with Qwen3-VL-8B-Instruct~\citep{bai2025qwen3vl} as the base model. The learning rate is set to $1\times10^{-5}$ with a cosine schedule, the batch size to 32, and the maximum context length to 40K tokens. For both trajectory collection and inference, the agent is limited to 8 tool calls, and each \texttt{image\_search} returns at most 5 images.

\providecolor{oursrow}{RGB}{240,246,252}

\begin{table}[t]
    \centering
    \caption{Search and generation results on \ourbench{} by domain. Bold/underline mark best/second-best scores. $^{*}$ indicates methods with a stronger model. Qwen-IE-2511 denotes Qwen-Image-Edit-2511.}
    \label{tab:evogen-main-original}
    \begingroup
    \footnotesize
    \setlength{\tabcolsep}{3pt}
    \renewcommand{\arraystretch}{1.15}
    \resizebox{\linewidth}{!}{%
    \begin{tabular}{@{}ll*{12}{c}@{}}
        \toprule
        \multirow{2.5}{*}{\textbf{Method}} & \multirow{2.5}{*}{\textbf{Generator}}
            & \multicolumn{2}{c}{\textbf{Entertainment}}
            & \multicolumn{2}{c}{\textbf{Sports}}
            & \multicolumn{2}{c}{\textbf{Fashion}}
            & \multicolumn{2}{c}{\textbf{Products}}
            & \multicolumn{2}{c}{\textbf{Culture}}
            & \multicolumn{2}{c}{\textbf{All}} \\
        \cmidrule(lr){3-4}\cmidrule(lr){5-6}\cmidrule(lr){7-8}
        \cmidrule(lr){9-10}\cmidrule(lr){11-12}\cmidrule(l){13-14}
        & & Search & Gen. & Search & Gen. & Search & Gen.
          & Search & Gen. & Search & Gen. & Search & Gen. \\
        \midrule
        \rowcolor{gray!15}
        \multicolumn{14}{c}{\textit{Proprietary Generative Systems}} \\
        -- & GPT-Image-2 & -- & 23.4 & -- & 29.5 & -- & 25.3 & -- & 29.9 & -- & 23.7 & -- & 26.4 \\
        -- & Nano Banana Pro & -- & 25.9 & -- & 30.5 & -- & 23.4 & -- & 28.5 & -- & 22.4 & -- & 26.1 \\
        -- & Nano Banana 2 & -- & 33.6 & -- & 37.7 & -- & 31.3 & -- & 39.2 & -- & 32.6 & -- & 34.9 \\
        \midrule
        \rowcolor{gray!15}
        \multicolumn{14}{c}{\textit{Harness Frameworks}} \\
        \multirow{2}{*}{\shortstack[l]{Mind-Brush\\{\scriptsize\textcolor{black!45}{GPT-5.1}}}} & Qwen-IE-2511 & \multirow{2}{*}{39.0} & 31.4 & \multirow{2}{*}{38.3} & 26.8 & \multirow{2}{*}{29.8} & 23.8 & \multirow{2}{*}{45.7} & 33.1 & \multirow{2}{*}{26.3} & 32.6 & \multirow{2}{*}{35.8} & 29.5 \\
         & Nano Banana Pro &  & 47.6 &  & 40.6 &  & 36.4 &  & 46.7 &  & 39.5 &  & 42.2 \\
        \midrule
        \multirow{2}{*}{\shortstack[l]{SCOPE\\{\scriptsize\textcolor{black!45}{GPT-5.4}}}} & Qwen-IE-2511 & \multirow{2}{*}{49.2} & 37.1 & \multirow{2}{*}{38.3} & 22.2 & \multirow{2}{*}{35.1} & 27.9 & \multirow{2}{*}{56.4} & 36.4 & \multirow{2}{*}{26.3} & 27.4 & \multirow{2}{*}{41.1} & 30.2 \\
         & Nano Banana Pro &  & \underline{49.4} &  & 42.7 &  & \underline{44.5} &  & \underline{50.8} &  & 38.0 &  & 45.1 \\
        \midrule

        \multirow{2}{*}{\shortstack[l]{SCOPE$^{*}$\\{\scriptsize\textcolor{black!45}{GPT-6-Astra}}}} & Qwen-IE-2511 & \multirow{2}{*}{\underline{55.9}} & 39.5 & \multirow{2}{*}{\underline{48.3}} & \underline{37.3} & \multirow{2}{*}{\underline{39.4}} & 30.1 & \multirow{2}{*}{\underline{59.6}} & \underline{39.9} & \multirow{2}{*}{\underline{54.4}} & 39.5 & \multirow{2}{*}{\underline{51.5}} & 37.3 \\
         & Nano Banana Pro &  & 47.3 &  & 48.5 &  & 41.3 &  & \underline{50.8} &  & 50.0 &  & \underline{47.6} \\
        \midrule
        \multirow{2}{*}{\shortstack[l]{GenClaw\\{\scriptsize\textcolor{black!45}{Claude-Opus-4.6}}}} & Qwen-IE-2511 & \multirow{2}{*}{51.8} & 37.7 & \multirow{2}{*}{40.0} & 33.6 & \multirow{2}{*}{36.2} & 28.9 & \multirow{2}{*}{55.3} & 37.2 & \multirow{2}{*}{42.1} & 33.9 & \multirow{2}{*}{45.1} & 34.2 \\
         & Nano Banana Pro &  & 46.4 &  & 45.8 &  & 41.2 &  & 47.0 &  & 40.9 &  & 44.2 \\
        \midrule
        \multirow{2}{*}{\shortstack[l]{GenClaw$^{*}$\\{\scriptsize\textcolor{black!45}{Claude-Opus-5}}}} & Qwen-IE-2511 & \multirow{2}{*}{55.4} & 39.3 & \multirow{2}{*}{43.3} & 35.1 & \multirow{2}{*}{38.3} & 29.7 & \multirow{2}{*}{\underline{59.6}} & 39.5 & \multirow{2}{*}{52.6} & 38.7 & \multirow{2}{*}{49.8} & 36.5 \\
         & Nano Banana Pro &  & 47.0 &  & 45.3 &  & 40.7 &  & 50.3 &  & 48.9 &  & 46.4 \\
        \midrule
        \rowcolor{gray!15}
        \multicolumn{14}{c}{\textit{Unified Multimodal Models}} \\
        \multirow{3}{*}{\shortstack[l]{Unify-Agent\\{\scriptsize\textcolor{black!45}{BAGEL-14B}}}} & BAGEL & \multirow{3}{*}{20.0} & 21.1 & \multirow{3}{*}{18.3} & 23.2 & \multirow{3}{*}{22.3} & 21.9 & \multirow{3}{*}{30.9} & 22.9 & \multirow{3}{*}{24.6} & 23.2 & \multirow{3}{*}{23.2} & 22.5 \\
         & Qwen-IE-2511 &  & 23.5 &  & 24.9 &  & 24.5 &  & 24.9 &  & 26.8 &  & 24.9 \\
         & Nano Banana Pro &  & 30.4 &  & 31.9 &  & 29.9 &  & 38.6 &  & 27.2 &  & 31.6 \\
        \midrule
        \rowcolor{gray!15}
        \multicolumn{14}{c}{\textit{Agentic Post-trained Models}} \\
        \multirow{2}{*}{\shortstack[l]{Gen-Searcher\\{\scriptsize\textcolor{black!45}{Qwen3-VL-8B}}}} & Qwen-IE-2511 & \multirow{2}{*}{46.7} & 35.5 & \multirow{2}{*}{36.7} & 36.1 & \multirow{2}{*}{36.2} & \underline{31.9} & \multirow{2}{*}{48.9} & 38.2 & \multirow{2}{*}{49.1} & \underline{47.7} & \multirow{2}{*}{43.5} & \underline{37.9} \\
         & Nano Banana Pro &  & 44.9 &  & \underline{48.7} &  & 44.1 &  & 45.8 &  & \underline{52.8} &  & 47.3 \\
        \midrule
        \multirow{2}{*}{\shortstack[l]{GenEvolve\\{\scriptsize\textcolor{black!45}{Qwen3-VL-8B}}}} & Qwen-IE-2511 & \multirow{2}{*}{45.1} & \underline{41.3} & \multirow{2}{*}{36.7} & 28.6 & \multirow{2}{*}{29.8} & 29.3 & \multirow{2}{*}{42.6} & 33.0 & \multirow{2}{*}{33.3} & 38.9 & \multirow{2}{*}{37.5} & 34.2 \\
         & Nano Banana Pro &  & 48.2 &  & 43.6 &  & 36.7 &  & 43.3 &  & 41.7 &  & 42.7 \\
        \midrule
        \rowcolor{oursrow}
         & Qwen-IE-2511 &  & \textbf{50.0} &  & \textbf{39.6} &  & \textbf{48.8} &  & \textbf{45.3} &  & \textbf{50.3} &  & \textbf{46.8} \\
        \rowcolor{oursrow}
        \multirow{-2}{*}{\shortstack[l]{\ourmodel{}\\{\scriptsize\textcolor{black!45}{Qwen3-VL-8B}}}} & Nano Banana Pro & \multirow{-2}{*}{\textbf{71.3}} & \textbf{58.6} & \multirow{-2}{*}{\textbf{66.7}} & \textbf{59.1} & \multirow{-2}{*}{\textbf{71.3}} & \textbf{65.5} & \multirow{-2}{*}{\textbf{61.7}} & \textbf{52.6} & \multirow{-2}{*}{\textbf{64.9}} & \textbf{58.7} & \multirow{-2}{*}{\textbf{67.2}} & \textbf{58.9} \\
        \bottomrule
    \end{tabular}%
    }
    \endgroup
    \vspace{-8pt}
\end{table}

\vspace{-6pt}
\section{Experiments}
\label{sec:exp}

\definecolor{oursrow}{RGB}{240,246,252}

\vspace{-6pt}
\subsection{Evaluation Setup}
\label{sec:exp_setup}

\vspace{-6pt}
\paragraph{Baselines.}
We compare \ourmodel{}-8B with four families of baselines:
(1) Proprietary generative systems, GPT-Image-2~\citep{openai2026gptimage2}, Nano Banana Pro~\citep{google2025nanobananapro},
and Nano Banana 2~\citep{google2026nanobanana2}, which we call through their official APIs with
built-in web search enabled;
(2) Harness frameworks, Mind-Brush~\citep{he2026mindbrush}, SCOPE~\citep{ren2026scope}, and GenClaw~\citep{ye2026genclaw},
which equip proprietary LLMs (GPT-5.1~\citep{openai2025gpt51}, GPT-5.4~\citep{openai2026gpt54},
and Claude-Opus-4.6~\citep{anthropic2026opus46}, respectively) with search tools, where SCOPE and GenClaw are
additionally run with the stronger GPT-6-Astra~\citep{openai2026gpt6astra} and Claude-Opus-5~\citep{anthropic2026opus5};
(3) Unified multimodal models, represented by Unify-Agent~\citep{chen2026unifyagent};
and (4) Agentic post-trained models, Gen-Searcher~\citep{feng2026gensearcher} and
GenEvolve~\citep{chen2026genevolve}, which share the same backbone with \ourmodel{}.

\vspace{-6pt}
\paragraph{Benchmarks.}
We mainly evaluate on \ourbench{} with two downstream generators, Qwen-Image-Edit-2511~\citep{wu2025qwenimage} and Nano Banana Pro~\citep{google2025nanobananapro}. To test generalization beyond \ourbench{}, we further evaluate on two representative benchmarks, Mind-Bench~\citep{he2026mindbrush} and KnowGen~\citep{feng2026gensearcher}. Benchmark and evaluation details are given in \cref{sec:app_metrics}. 

\begin{table}[t]
    \centering
    \caption{Results on Mind-Bench and KnowGen.
    Bold/underline mark best/second-best scores
    among open-source methods.}
    \label{tab:mindbench-knowgen-baselines}
    \begingroup
    \footnotesize
    \setlength{\tabcolsep}{3pt}
    \renewcommand{\arraystretch}{1.15}
    \resizebox{\linewidth}{!}{%
    \begin{tabular}{@{}ll*{8}{c}@{}}
        \toprule
        \multirow{2.5}{*}{\textbf{Method}} & \multirow{2.5}{*}{\textbf{Generator}}
            & \multicolumn{3}{c}{\textbf{Mind-Bench}}
            & \multicolumn{5}{c}{\textbf{KnowGen}} \\
        \cmidrule(lr){3-5}\cmidrule(l){6-10}
        & & Knowledge & Reasoning& Overall
            & Visual cor. & Text acc. & Faith. & Aesth. & K-Score \\
        \midrule
        \rowcolor{gray!15}
        \multicolumn{10}{c}{\textit{Proprietary Generative Systems}} \\
        -- & GPT-Image-1
            & 15.6 & 17.6 & 17.0 & 20.18 & 29.94 & 78.72 & 62.78 & 34.19 \\
        -- & GPT-Image-1.5
            & 22.0 & 17.6 & 21.0 & 29.34 & 43.18 & 85.47 & 74.19 & 44.97 \\
        -- & Nano Banana
            & 16.4 & 20.0 & 18.0 & 16.14 & 22.72 & 78.59 & 68.37 & 30.24 \\
        -- & Nano Banana Pro
            & 39.6 & 44.4 & 41.0 & 34.99 & 51.35 & 88.65 & 69.84 & 50.38 \\
        \midrule
        \rowcolor{gray!15}
        \multicolumn{10}{c}{\textit{Open-source Generative Models}} \\
        -- & BAGEL
            & 0.4 & 2.8 & 2.0 & 6.63 & 2.15 & 53.76 & \underline{52.72} & 13.85 \\
        -- & Z-Image
            & 2.4 & 0.0 & 2.0 & 7.35 & 2.36 & 55.70 & 50.42 & 14.49 \\
        -- & Qwen-Image
            & 2.4 & 1.6 & 2.0 & 7.20 & 0.87 & 58.18 & \textbf{59.35} & 14.98 \\
        \midrule
        \rowcolor{gray!15}
        \multicolumn{10}{c}{\textit{Agentic Image Generation Methods}} \\
        Mind-Brush & Qwen-IE-2511
            & 25.2 & 6.4 & 15.8 & 17.42 & 1.66 & 49.16 & 41.88 & 16.74 \\
        Gen-Searcher & Qwen-IE-2511
            & \underline{27.6} & 12.0 & \underline{19.8} & \underline{23.92} & \textbf{14.21} & \underline{65.86} & 45.22 & \underline{26.35} \\
        GenEvolve & Qwen-IE-2511
            & 20.8 & \underline{16.0} & 18.4 & 15.75 & \underline{13.65} & 63.32 & 48.16 & 22.91 \\
        \rowcolor{oursrow}
        \ourmodel{} & Qwen-IE-2511
            & \textbf{34.0} & \textbf{22.4} & \textbf{28.2} & \textbf{26.20} & 13.19 & \textbf{68.70} & 51.25 & \textbf{27.75} \\
        \bottomrule
    \end{tabular}%
    }
    \par\smallskip
    \endgroup
\end{table}

\subsection{Main Results on \ourbench{}}
\label{sec:exp_main}

\paragraph{Existing methods struggle in the real-world setting.}
As shown in \cref{tab:evogen-main-original}, no baseline finds the right reference reliably. Even harness frameworks equipped with the latest frontier LLMs (\eg SCOPE with GPT-6-Astra and GenClaw with Claude-Opus-5) pass only about half of the prompts. The difficulty is sharpest in Fashion, where no baseline exceeds 40. Proprietary generation systems do not escape it with their built-in search either, scoring at most 34.9 on generation. These results show that existing methods are not yet ready for image generation in the wild, where the target must be found on a crowded and evolving web.

\paragraph{Generation is bounded by reference quality.}
A strong generator cannot compensate for incorrect references. Unify-Agent passes the same references to three generators, and even Nano Banana Pro lifts its score only to 31.6. Conversely, the same Nano Banana Pro scores 26.1 with its own search but 58.9 with references from \ourmodel{}. Gains in generation also follow gains in search: Fashion, where \ourmodel{} improves search the most, is also where its generation gains are largest, 16.9 and 21.0 points with the two generators.

\paragraph{\ourmodel{} leads the way.}
With only 8B parameters, \textbf{\ourmodel{} outperforms harness frameworks built on GPT-6-Astra and Claude-Opus-5}, raising the search score from 51.5 to 67.2 and achieving the best generation score in every domain with both generators. Its references are good enough that the open-source Qwen-Image-Edit-2511 (46.8) nearly matches the best baseline running on Nano Banana Pro (47.6). Compared with Gen-Searcher and GenEvolve, which share the same backbone, \ourmodel{} leads by more than 20 search points, demonstrating the effectiveness of the proposed paradigm. Comparisons on trajectories, generation quality and additional case studies are presented in \cref{sec:app_trajectories,sec:app_cases_main,sec:app_generation_gallery}.

\subsection{Generalization Performance}
\label{sec:exp_transfer}

Although trained only on news-driven data, \textbf{\ourmodel{} achieves the best overall score among agentic methods on both benchmarks} (\cref{tab:mindbench-knowgen-baselines}). Notably, these are the benchmarks proposed with Mind-Brush and Gen-Searcher, respectively, and \ourmodel{} outperforms each on its own benchmark, with 28.2 vs.\ 15.8 on Mind-Bench and 27.75 vs.\ 26.35 on KnowGen.

On Mind-Bench, \ourmodel{} brings clear gains on knowledge categories that hinge on a specific event or character, such as Special Events and New Movie Characters. On KnowGen, it stands out in visual correctness, which checks whether the depicted entity matches the ground truth and thus depends directly on the references, and it also leads in faithfulness to the prompt. These results show that the retrieval ability of \ourmodel{} carries over to benchmarks beyond \ourbench{}.

\subsection{Further Analysis}
\label{sec:exp_analysis}
 
We examine whether each phase of the proposed paradigm addresses the challenge it targets. We also validate the reliability of the date filtering method in \cref{sec:app_time_window} and include a human-model alignment evaluation in \cref{sec:app_human_agreement}.

\begin{figure}[!htb]
    \centering
    \captionsetup{font=small,skip=4pt}
    \begin{minipage}[t]{0.49\linewidth}
        \centering
        \includegraphics[width=\linewidth]{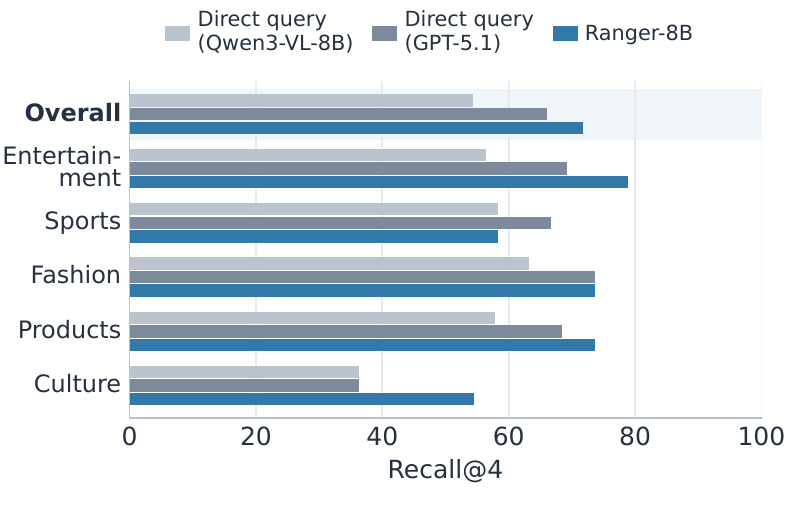}
        \caption{Comparison of Recall@4 of the first image search of direct query and Ranger.}
        \label{fig:locate-recall}
    \end{minipage}\hfill
    \begin{minipage}[t]{0.49\linewidth}
        \centering
        \includegraphics[width=\linewidth]{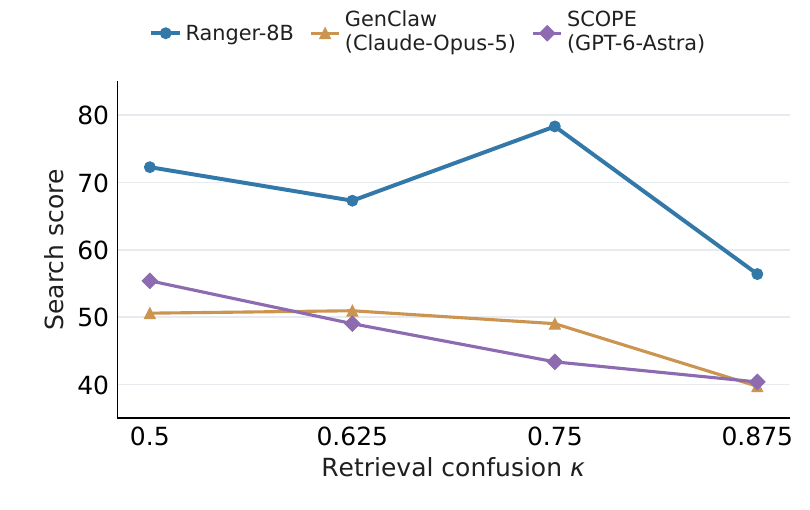}
        \caption{Search score grouped by retrieval confusion.}
        \label{fig:corroborate-confusion}
    \end{minipage}
\end{figure}

\paragraph{Locate recovers targets that direct search misses.}
We compare the first image search of \ourmodel{} with direct image searches directly derived from $p$ by Qwen3-VL-8B and GPT-5.1 without text search. As shown in \cref{fig:locate-recall}, \ourmodel{} reaches a Recall@4 of 71.7, compared with 54.4 and 66.0 for the two direct baselines. It proves that gathering textual evidence first makes the image query specific enough to find a newly introduced target, which addresses temporal novelty.

\paragraph{Corroborate resists semantic crowding.}
We group the 500 prompts by retrieval confusion $\kappa$ (\cref{eq:retrieval_confusion}) and compare \ourmodel{} with GenClaw (Claude-Opus-5) and SCOPE (GPT-6-Astra). As shown in \cref{fig:corroborate-confusion}, \ourmodel{} leads by at least 16 points in every group. In the most confusing group, it reaches a search score of 56.4, compared with 39.7 and 40.4 for the two baselines. Checking candidates across modalities and across images thus keeps the agent on the right target even when most search results show other entities, which addresses semantic crowding.

\par\addvspace{\medskipamount}
\noindent\begin{minipage}{\textwidth}
\begin{minipage}[t]{0.48\linewidth}
\vspace{0pt}
\paragraph{Isolate improves generation quality.}
On 102 prompts where \ourmodel{} calls \texttt{crop\_image}, we compare the images generated from the cropped reference with those generated from the full source image. As shown in \cref{fig:isolate-crop}, cropping raises the identity score by 6.3 points with Qwen-Image-Edit-2511 and 3.8 points with Nano Banana Pro, while the context score improves less, by 1.4 and 1.3 points. These results show that removing the surrounding content helps the generator depict the right entity without losing the requested context, which addresses spatial entanglement. Generation examples are given in \cref{sec:app_crop_cases}.
\end{minipage}\hfill
\begin{minipage}[t]{0.49\linewidth}
    \vspace{0pt}
    \centering
    \includegraphics[width=\linewidth]{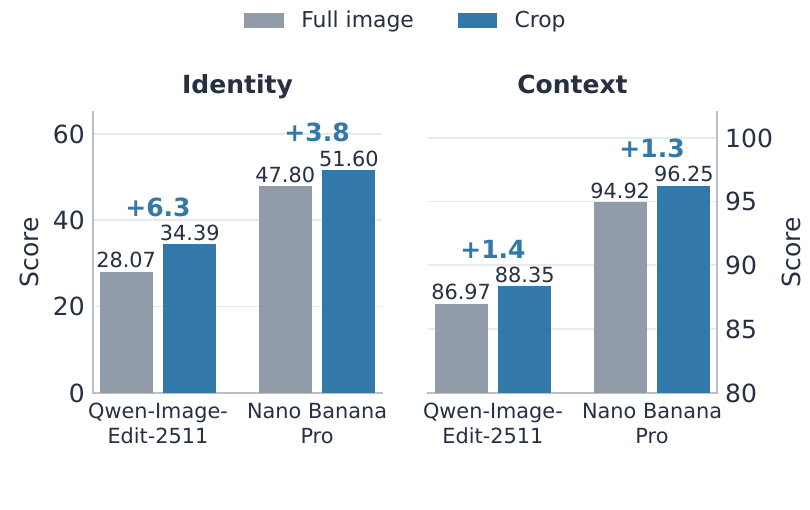}
    \captionsetup{font=small,skip=4pt}
    \captionof{figure}{Generation scores with full and cropped references. The context axis starts at 80.}
    \label{fig:isolate-crop}
\end{minipage}
\end{minipage}
\par\addvspace{\medskipamount}

\section{Conclusion}
\label{sec:conclusion}
We studied agentic image generation in the wild and identified three challenges it poses: temporal novelty, semantic crowding, and spatial entanglement. We built a news-driven pipeline that reproduces these challenges under real search conditions, and used it to construct \ourbench{}, on which even harness frameworks built on frontier models find a correct reference for only about half of the cases. We then proposed the Locate-Corroborate-Isolate paradigm. Trained on trajectories that follow it, \ourmodel{}-8B raises the search score from 51.5 to 67.2, gives the best generation quality with every generator we test, and surpasses prior agentic methods on other representative benchmarks. Our analysis further confirms that each phase addresses its intended challenge.

\bibliography{refs}

\begin{thebibliography}{35}
\providecommand{\natexlab}[1]{#1}
\providecommand{\url}[1]{\texttt{#1}}
\expandafter\ifx\csname urlstyle\endcsname\relax
  \providecommand{\doi}[1]{doi: #1}\else
  \providecommand{\doi}{doi: \begingroup \urlstyle{rm}\Url}\fi

\bibitem[{Anthropic}(2026{\natexlab{a}})]{anthropic2026opus46}
{Anthropic}.
\newblock Introducing {Claude Opus} 4.6.
\newblock \url{https://www.anthropic.com/news/claude-opus-4-6}, 2026{\natexlab{a}}.

\bibitem[{Anthropic}(2026{\natexlab{b}})]{anthropic2026opus5}
{Anthropic}.
\newblock Introducing {Claude Opus} 5.
\newblock \url{https://www.anthropic.com/news/claude-opus-5}, 2026{\natexlab{b}}.

\bibitem[Bai et~al.(2025)Bai, Cai, Chen, Chen, Chen, Cheng, Deng, Ding, Gao, Ge, Ge, Guo, Huang, Huang, Huang, Hui, Jiang, Li, Li, Li, Li, Lin, Lin, Liu, Liu, Liu, Liu, Liu, Liu, Lu, Luo, Lv, Men, Meng, Ren, Ren, Song, Sun, Tang, Tu, Wan, Wang, Wang, Wang, Wang, Xie, Xu, Xu, Xu, Yang, Yang, Yang, Yang, Yu, Zhang, Zhang, Zhang, Zheng, Zhong, Zhou, Zhou, Zhou, Zhu, and Zhu]{bai2025qwen3vl}
Shuai Bai, Yuxuan Cai, Ruizhe Chen, Keqin Chen, Xionghui Chen, Zesen Cheng, Lianghao Deng, Wei Ding, Chang Gao, Chunjiang Ge, Wenbin Ge, Zhifang Guo, Qidong Huang, Jie Huang, Fei Huang, Binyuan Hui, Shutong Jiang, Zhaohai Li, Mingsheng Li, Mei Li, Kaixin Li, Zicheng Lin, Junyang Lin, Xuejing Liu, Jiawei Liu, Chenglong Liu, Yang Liu, Dayiheng Liu, Shixuan Liu, Dunjie Lu, Ruilin Luo, Chenxu Lv, Rui Men, Lingchen Meng, Xuancheng Ren, Xingzhang Ren, Sibo Song, Yuchong Sun, Jun Tang, Jianhong Tu, Jianqiang Wan, Peng Wang, Pengfei Wang, Qiuyue Wang, Yuxuan Wang, Tianbao Xie, Yiheng Xu, Haiyang Xu, Jin Xu, Zhibo Yang, Mingkun Yang, Jianxin Yang, An~Yang, Bowen Yu, Fei Zhang, Hang Zhang, Xi~Zhang, Bo~Zheng, Humen Zhong, Jingren Zhou, Fan Zhou, Jing Zhou, Yuanzhi Zhu, and Ke~Zhu.
\newblock Qwen3-vl technical report.
\newblock \emph{arXiv preprint arXiv:2511.21631}, 2025.

\bibitem[Chen et~al.(2026{\natexlab{a}})Chen, Shou, Chen, Zhou, Feng, Hu, Zhang, Lin, Huang, Song, Dai, Jiang, Zhang, Zhang, Jiang, Wang, Zhong, Cheng, and Peng]{chen2026unifyagent}
Shuang Chen, Quanxin Shou, Hangting Chen, Yucheng Zhou, Kaituo Feng, Wenbo Hu, Yi{-}Fan Zhang, Yunlong Lin, Wenxuan Huang, Mingyang Song, Dasen Dai, Bolin Jiang, Manyuan Zhang, Shi{-}Xue Zhang, Zhengkai Jiang, Lucas Wang, Zhao Zhong, Yu~Cheng, and Nanyun Peng.
\newblock Unify-agent: {A} unified multimodal agent for world-grounded image synthesis.
\newblock \emph{CoRR}, abs/2603.29620, 2026{\natexlab{a}}.

\bibitem[Chen et~al.(2026{\natexlab{b}})Chen, Xing, Ye, Geng, Lin, Lai, He, Zhai, Gao, and Zhu]{chen2026genevolve}
Sixiang Chen, Zhaohu Xing, Tian Ye, Xinyu Geng, Yunlong Lin, Jianyu Lai, Xuanhua He, Fuxiang Zhai, Jialin Gao, and Lei Zhu.
\newblock Genevolve: Self-evolving image generation agents via tool-orchestrated visual experience distillation.
\newblock \emph{CoRR}, abs/2605.21605, 2026{\natexlab{b}}.

\bibitem[Deng et~al.(2025)Deng, Zhu, Li, Gou, Li, Wang, Zhong, Yu, Nie, Song, Shi, and Fan]{deng2025bagel}
Chaorui Deng, Deyao Zhu, Kunchang Li, Chenhui Gou, Feng Li, Zeyu Wang, Shu Zhong, Weihao Yu, Xiaonan Nie, Ziang Song, Guang Shi, and Haoqi Fan.
\newblock Emerging properties in unified multimodal pretraining.
\newblock \emph{arXiv preprint arXiv:2505.14683}, 2025.

\bibitem[Du et~al.(2026)Du, Liu, Peng, Wu, Li, Li, Zhao, and Wen]{du2026longhorizon}
Yifan Du, Zikang Liu, Jinbiao Peng, Jie Wu, Junyi Li, Jinyang Li, Wayne~Xin Zhao, and Ji{-}Rong Wen.
\newblock Towards long-horizon agentic multimodal search.
\newblock \emph{CoRR}, abs/2604.12890, 2026.

\bibitem[Esser et~al.(2024)Esser, Kulal, Blattmann, Entezari, M{\"{u}}ller, Saini, Levi, Lorenz, Sauer, Boesel, Podell, Dockhorn, English, and Rombach]{esser2024sd3}
Patrick Esser, Sumith Kulal, Andreas Blattmann, Rahim Entezari, Jonas M{\"{u}}ller, Harry Saini, Yam Levi, Dominik Lorenz, Axel Sauer, Frederic Boesel, Dustin Podell, Tim Dockhorn, Zion English, and Robin Rombach.
\newblock Scaling rectified flow transformers for high-resolution image synthesis.
\newblock In \emph{{ICML}}, volume 235 of \emph{Proceedings of Machine Learning Research}, pp.\  12606--12633. {PMLR} / OpenReview.net, 2024.

\bibitem[Feng et~al.(2026)Feng, Zhang, Chen, Lin, Fan, Jiang, Li, Zheng, Wang, and Yue]{feng2026gensearcher}
Kaituo Feng, Manyuan Zhang, Shuang Chen, Yunlong Lin, Kaixuan Fan, Yilei Jiang, Hongyu Li, Dian Zheng, Chenyang Wang, and Xiangyu Yue.
\newblock Gen-searcher: Reinforcing agentic search for image generation.
\newblock \emph{CoRR}, abs/2603.28767, 2026.

\bibitem[Geng et~al.(2025)Geng, Xia, Zhang, Wang, Wang, Ding, Wang, Wu, Zhao, Li, Jiang, Xie, Huang, and Zhou]{geng2025webwatcher}
Xinyu Geng, Peng Xia, Zhen Zhang, Xinyu Wang, Qiuchen Wang, Ruixue Ding, Chenxi Wang, Jialong Wu, Yida Zhao, Kuan Li, Yong Jiang, Pengjun Xie, Fei Huang, and Jingren Zhou.
\newblock Webwatcher: Breaking new frontier of vision-language deep research agent.
\newblock \emph{CoRR}, abs/2508.05748, 2025.

\bibitem[{Google}(2025{\natexlab{a}})]{google2025nanobanana}
{Google}.
\newblock Introducing {Gemini} 2.5 {Flash} {Image}, our state-of-the-art image model.
\newblock \url{https://developers.googleblog.com/introducing-gemini-2-5-flash-image/}, 2025{\natexlab{a}}.

\bibitem[{Google}(2025{\natexlab{b}})]{google2025nanobananapro}
{Google}.
\newblock Introducing nano banana pro.
\newblock \url{https://blog.google/innovation-and-ai/products/nano-banana-pro/}, 2025{\natexlab{b}}.

\bibitem[{Google}(2026{\natexlab{a}})]{google2026gemini31pro}
{Google}.
\newblock {Gemini} 3.1 {Pro}: A smarter model for your most complex tasks.
\newblock \url{https://blog.google/innovation-and-ai/models-and-research/gemini-models/gemini-3-1-pro/}, 2026{\natexlab{a}}.

\bibitem[{Google}(2026{\natexlab{b}})]{google2026nanobanana2}
{Google}.
\newblock {Nano Banana} 2: Combining {Pro} capabilities with lightning-fast speed.
\newblock \url{https://blog.google/innovation-and-ai/technology/ai/nano-banana-2/}, 2026{\natexlab{b}}.

\bibitem[He et~al.(2026)He, Ye, Huang, Jiang, Zhang, Zhu, Zhang, Zhang, and Li]{he2026mindbrush}
Jun He, Junyan Ye, Zilong Huang, Dongzhi Jiang, Chenjue Zhang, Leqi Zhu, Renrui Zhang, Xiang Zhang, and Weijia Li.
\newblock Mind-brush: Integrating agentic cognitive search and reasoning into image generation.
\newblock \emph{CoRR}, abs/2602.01756, 2026.

\bibitem[Li et~al.(2025{\natexlab{a}})Li, Sun, Feng, Zhai, Chang, and Zhang]{li2025iat2i}
Chuanhao Li, Jianwen Sun, Yukang Feng, Mingliang Zhai, Yifan Chang, and Kaipeng Zhang.
\newblock {IA-T2I:} internet-augmented text-to-image generation.
\newblock \emph{CoRR}, abs/2505.15779, 2025{\natexlab{a}}.

\bibitem[Li et~al.(2025{\natexlab{b}})Li, Jin, Dong, Qian, Wu, Wen, Zhu, and Dou]{li2025webthinker}
Xiaoxi Li, Jiajie Jin, Guanting Dong, Hongjin Qian, Yongkang Wu, Ji{-}Rong Wen, Yutao Zhu, and Zhicheng Dou.
\newblock Webthinker: Empowering large reasoning models with deep research capability.
\newblock In \emph{NeurIPS}, 2025{\natexlab{b}}.

\bibitem[Nakano et~al.(2021)Nakano, Hilton, Balaji, Wu, Ouyang, Kim, Hesse, Jain, Kosaraju, Saunders, Jiang, Cobbe, Eloundou, Krueger, Button, Knight, Chess, and Schulman]{nakano2021webgpt}
Reiichiro Nakano, Jacob Hilton, Suchir Balaji, Jeff Wu, Long Ouyang, Christina Kim, Christopher Hesse, Shantanu Jain, Vineet Kosaraju, William Saunders, Xu~Jiang, Karl Cobbe, Tyna Eloundou, Gretchen Krueger, Kevin Button, Matthew Knight, Benjamin Chess, and John Schulman.
\newblock Webgpt: Browser-assisted question-answering with human feedback.
\newblock \emph{CoRR}, abs/2112.09332, 2021.

\bibitem[Niu et~al.(2025)Niu, Ning, Zheng, Lin, Jin, Liao, Ning, Zhu, and Yuan]{niu2026wise}
Yuwei Niu, Munan Ning, Mengren Zheng, Bin Lin, Peng Jin, Jiaqi Liao, Kun{-}Peng Ning, Bin Zhu, and Li~Yuan.
\newblock {WISE:} {A} world knowledge-informed semantic evaluation for text-to-image generation.
\newblock \emph{CoRR}, abs/2503.07265, 2025.

\bibitem[{OpenAI}(2025{\natexlab{a}})]{openai2025gpt51}
{OpenAI}.
\newblock {GPT-5.1}: A smarter, more conversational {ChatGPT}.
\newblock \url{https://openai.com/index/gpt-5-1/}, 2025{\natexlab{a}}.

\bibitem[{OpenAI}(2025{\natexlab{b}})]{openai2025gptimage1}
{OpenAI}.
\newblock Introducing our latest image generation model in the {API}.
\newblock \url{https://openai.com/index/image-generation-api/}, 2025{\natexlab{b}}.

\bibitem[{OpenAI}(2025{\natexlab{c}})]{openai2025gptimage15}
{OpenAI}.
\newblock The new {ChatGPT} images is here.
\newblock \url{https://openai.com/index/new-chatgpt-images-is-here/}, 2025{\natexlab{c}}.

\bibitem[{OpenAI}(2026{\natexlab{a}})]{openai2026gpt54}
{OpenAI}.
\newblock Introducing {GPT-5.4}.
\newblock \url{https://openai.com/index/introducing-gpt-5-4/}, 2026{\natexlab{a}}.

\bibitem[{OpenAI}(2026{\natexlab{b}})]{openai2026gpt6astra}
{OpenAI}.
\newblock {GPT-6 Astra}: A new generation of intelligence.
\newblock \url{https://openai.com/index/gpt-6-astra/}, 2026{\natexlab{b}}.

\bibitem[{OpenAI}(2026{\natexlab{c}})]{openai2026gptimage2}
{OpenAI}.
\newblock Introducing {ChatGPT} images 2.0.
\newblock \url{https://openai.com/index/introducing-chatgpt-images-2-0/}, 2026{\natexlab{c}}.

\bibitem[Peebles \& Xie(2023)Peebles and Xie]{peebles2023dit}
William Peebles and Saining Xie.
\newblock Scalable diffusion models with transformers.
\newblock In \emph{{ICCV}}, pp.\  4172--4182. {IEEE}, 2023.

\bibitem[Ren et~al.(2026)Ren, Yan, Zhao, Fang, Zeng, Zhang, Xu, Ma, Huang, Xu, Huang, Wang, Chen, Chen, Huang, and Zhao]{ren2026scope}
Tianfei Ren, Zhipeng Yan, Yiming Zhao, Zhen Fang, Yu~Zeng, Guohui Zhang, Hang Xu, Xiaoxiao Ma, Shiting Huang, Ke~Xu, Wenxuan Huang, Lionel~Z. Wang, Lin Chen, Zehui Chen, Jie Huang, and Feng Zhao.
\newblock {SCOPE:} structured decomposition and conditional skill orchestration for complex image generation.
\newblock \emph{CoRR}, abs/2605.08043, 2026.

\bibitem[Rombach et~al.(2022)Rombach, Blattmann, Lorenz, Esser, and Ommer]{rombach2022ldm}
Robin Rombach, Andreas Blattmann, Dominik Lorenz, Patrick Esser, and Bj{\"{o}}rn Ommer.
\newblock High-resolution image synthesis with latent diffusion models.
\newblock In \emph{{CVPR}}, pp.\  10674--10685. {IEEE}, 2022.

\bibitem[Team et~al.(2025)Team, Cai, Cao, Du, Gao, Hoi, Hou, Huang, Jiang, Jin, Li, Li, Li, Liu, Liu, Shi, Wu, Yu, Zhang, Zhang, and Zhou]{cai2025zimage}
Z{-}Image Team, Huanqia Cai, Sihan Cao, Ruoyi Du, Peng Gao, Steven C.~H. Hoi, Zhaohui Hou, Shijie Huang, Dengyang Jiang, Xin Jin, Liangchen Li, Zhen Li, Zhong{-}Yu Li, David Liu, Dongyang Liu, Junhan Shi, Qilong Wu, Feng Yu, Chi Zhang, Shifeng Zhang, and Shilin Zhou.
\newblock Z-image: An efficient image generation foundation model with single-stream diffusion transformer.
\newblock \emph{CoRR}, abs/2511.22699, 2025.

\bibitem[Wei et~al.(2025)Wei, Sun, Papay, McKinney, Han, Fulford, Chung, Passos, Fedus, and Glaese]{wei2025browsecomp}
Jason Wei, Zhiqing Sun, Spencer Papay, Scott McKinney, Jeffrey Han, Isa Fulford, Hyung~Won Chung, Alex~Tachard Passos, William Fedus, and Amelia Glaese.
\newblock Browsecomp: {A} simple yet challenging benchmark for browsing agents.
\newblock \emph{CoRR}, abs/2504.12516, 2025.

\bibitem[Wu et~al.(2025)Wu, Li, Zhou, Lin, Gao, Yan, Yin, Bai, Xu, Chen, Chen, Tang, Zhang, Wang, Yang, Yu, Cheng, Liu, Li, Zhang, Meng, Wei, Ni, Chen, Cao, Peng, Qu, Wu, Wang, Yu, Wen, Feng, Xu, Wang, Zhang, Zhu, Wu, Cai, and Liu]{wu2025qwenimage}
Chenfei Wu, Jiahao Li, Jingren Zhou, Junyang Lin, Kaiyuan Gao, Kun Yan, Shengming Yin, Shuai Bai, Xiao Xu, Yilei Chen, Yuxiang Chen, Zecheng Tang, Zekai Zhang, Zhengyi Wang, An~Yang, Bowen Yu, Chen Cheng, Dayiheng Liu, Deqing Li, Hang Zhang, Hao Meng, Hu~Wei, Jingyuan Ni, Kai Chen, Kuan Cao, Liang Peng, Lin Qu, Minggang Wu, Peng Wang, Shuting Yu, Tingkun Wen, Wensen Feng, Xiaoxiao Xu, Yi~Wang, Yichang Zhang, Yongqiang Zhu, Yujia Wu, Yuxuan Cai, and Zenan Liu.
\newblock Qwen-image technical report.
\newblock \emph{CoRR}, abs/2508.02324, 2025.

\bibitem[Wu et~al.(2026)Wu, Deng, Li, Liu, You, Li, Ma, and Liu]{wu2025mmsearchr1}
Jinming Wu, Zihao Deng, Wei Li, Yiding Liu, Bo~You, Bo~Li, Zejun Ma, and Ziwei Liu.
\newblock Mmsearch-r1: Incentivizing lmms to search.
\newblock In \emph{{ACL} {(1)}}, pp.\  2456--2487. Association for Computational Linguistics, 2026.

\bibitem[Yao et~al.(2023)Yao, Zhao, Yu, Du, Shafran, Narasimhan, and Cao]{yao2023react}
Shunyu Yao, Jeffrey Zhao, Dian Yu, Nan Du, Izhak Shafran, Karthik~R. Narasimhan, and Yuan Cao.
\newblock React: Synergizing reasoning and acting in language models.
\newblock In \emph{{ICLR}}. OpenReview.net, 2023.

\bibitem[Ye et~al.(2026)Ye, He, Huang, Jiang, Yang, Chen, and Li]{ye2026genclaw}
Junyan Ye, Jun He, Zilong Huang, Dongzhi Jiang, Xuan Yang, Rui Chen, and Weijia Li.
\newblock Genclaw: Code-driven agentic image generation.
\newblock \emph{CoRR}, abs/2605.30248, 2026.

\bibitem[Zheng et~al.(2025)Zheng, Yang, Hong, Zhao, Xu, Yang, Shen, and Yu]{zheng2025deepeyes}
Ziwei Zheng, Michael Yang, Jack Hong, Chenxiao Zhao, Guohai Xu, Le~Yang, Chao Shen, and Xing Yu.
\newblock Deepeyes: Incentivizing "thinking with images" via reinforcement learning.
\newblock \emph{CoRR}, abs/2505.14362, 2025.

\end{thebibliography}
\bibliographystyle{ranger}

\appendix
\crefalias{section}{appendix}
\section{Related Work}
\label{sec:related}

\paragraph{Generative models.}
Diffusion and unified multimodal models~\citep{rombach2022ldm,peebles2023dit,esser2024sd3,deng2025bagel,cai2025zimage} now produce high-quality images, and recent generators~\citep{wu2025qwenimage,openai2025gptimage1,openai2025gptimage15,google2025nanobanana,google2025nanobananapro} also accept reference images as conditions. Their knowledge, however, is bounded by training data, so they fail on uncommon entities~\citep{niu2026wise}. Conditioning on references of the entity fills this gap, which leaves open where the references come from. Subject-driven methods assume the user provides them, and retrieval-augmented methods draw them from a fixed database that cannot cover entities introduced after it was built. We instead obtain references from the live web and keep the generator fixed.

\paragraph{Multimodal search agents.}
Search agents began with language models that interleave reasoning with web search~\citep{nakano2021webgpt,yao2023react}, and were later trained with reinforcement learning to search over many turns~\citep{li2025webthinker}, often on synthesized questions whose answers hide behind obscured clues~\citep{wei2025browsecomp}. Multimodal agents then added image search~\citep{wu2025mmsearchr1,geng2025webwatcher,du2026longhorizon}, and vision-language models learned to crop image regions for closer inspection~\citep{zheng2025deepeyes}. All of these agents end with a text answer, so a retrieved image only needs to be understood. An agent for generation must hand the image itself to the generator, so the image has to show the right entity and little else. \ourmodel{} therefore crops to prepare the generator's input rather than to inspect, and our query synthesis draws difficulty from real search confusion rather than from obscured clues.

\paragraph{Agentic image generation.}
Agentic image generation places an agent before the generator. Early agents only reason about and rewrite the prompt, so they remain limited to what the model knows. Later agents add web search to fetch text and images, either as training-free harnesses around frontier models~\citep{li2025iat2i,he2026mindbrush,ren2026scope,ye2026genclaw} or as trained agents~\citep{feng2026gensearcher,chen2026genevolve,chen2026unifyagent}. Their benchmarks~\citep{niu2026wise,he2026mindbrush,feng2026gensearcher} mostly ask for entities with an established visual record, where the difficulty lies in knowing what to search for. For newly introduced entities, the harder part comes after search, namely telling the target apart from similar entities and isolating it into a clean reference. \ourbench{} is built to test these two steps, and \ourmodel{} handles them in its Corroborate and Isolate phases.

\section{Tool Implementations}
\label{sec:app_tools}
\Cref{tab:tool_implementations} summarizes the tools used in our agent environment and the news retrieval service used for data construction. Serper provides the search backends; the agent receives retrieved text or image pixels through the corresponding tool interface.

\paragraph{News dates and retrieval time windows.}
We collect news day by day through Serper's \texttt{/news} endpoint, setting both \texttt{cd\_min} and \texttt{cd\_max} in the \texttt{tbs} parameter to the queried day, which we record as the news date $\tau$. For subsequent text and image searches, we set \texttt{tbs=cdr:1} with \texttt{cd\_min} set to January 1, 2000 and \texttt{cd\_max} set to $\tau+3$ days. When returned results include timestamps, such as \texttt{date} or \texttt{datePublished}, we also parse these fields and remove results with unambiguous dates later than the cutoff. Results without a usable, unambiguous date are retained.

\begin{table}[H]
\centering
\small
\setlength{\tabcolsep}{6pt}
\renewcommand{\arraystretch}{1.15}
\caption{Tool implementations. News retrieval is used during data construction; the remaining rows describe tools available to the agent.}
\label{tab:tool_implementations}
\begin{tabular}{@{}p{0.21\linewidth}p{0.23\linewidth}p{0.50\linewidth}@{}}
\toprule
Tool & Implementation & Inputs and returned evidence \\
\midrule
News retrieval & Serper \texttt{/news} & Queries a meta query within a single-day date window and returns news titles, snippets, source URLs, and publisher names for data construction. \\
\addlinespace
\texttt{text\_search} & Serper \texttt{/search} & Accepts one to three queries and returns ranked organic results with titles, snippets, and URLs; the default is five results per query. \\
\addlinespace
\texttt{image\_search} & Serper \texttt{/images} & Searches by a text query, downloads candidate images, and displays up to five images with image IDs, titles, source-page URLs, and image URLs. \\
\addlinespace
\texttt{web\_fetch} & Internal tool & Retrieves webpage content from a URL according to a supplied prompt. \\
\addlinespace
\texttt{crop\_image} & Local image operation & Takes a retrieved image ID and a bounding box in normalized $[0,1000]$ coordinates, then returns the cropped image with a new ID. \\
\bottomrule
\end{tabular}
\end{table}

\clearpage
\section{Meta Queries}
\label{sec:app_meta_query}
\Cref{tab:meta_queries} lists all 50 meta queries used to collect news for \ourbench{}. Each meta query names a family of visual artifacts within its subcategory, without pointing to a specific answer.

\begin{table}[h]
\centering
\small
\caption{Meta queries for each domain and subcategory of \ourbench{}.}
\label{tab:meta_queries}
\begin{tabular}{llp{0.58\linewidth}}
\toprule
Domain & Subcategory & Meta queries \\
\midrule
\multirow{5}{*}{Entertainment}
 & Film \& TV & movie costume first look; television series costume debut; film set design first look; movie poster artwork unveiling \\
 & Anime & anime key visual release; anime character design reveal \\
 & Games & video game character costume reveal; video game vehicle model debut; video game expansion content announcement \\
 & Collectibles & collectible figure reveal; comic issue variant cover reveal; collector edition box set reveal \\
 & Music & album cover artwork reveal; concert stage design debut; music video costume design \\
\midrule
\multirow{2}{*}{Sports}
 & General Sports & team kit design unveiling; championship trophy presentation; championship medal ceremony; player signing jersey presentation \\
 & Motorsport & race car livery unveiling; motorsport driver team announcement; grand prix podium trophy; motorsport driver equipment debut \\
\midrule
\multirow{3}{*}{Fashion}
 & Celebrities & celebrity press event appearance; red carpet gala arrival; celebrity street style outing \\
 & Collections & fashion week runway collection; luxury accessory collection launch \\
 & Watches \& Jewelry & watch model product unveiling; high jewelry collection unveiling \\
\midrule
\multirow{4}{*}{Products}
 & Tech & smartphone model product launch; wearable device product announcement; laptop computer product launch; gaming hardware product unveiling \\
 & Automobiles & production car model debut; concept vehicle design unveiling; prototype camouflage road testing; vehicle cabin interior reveal \\
 & Aviation \& Space & airline aircraft livery unveiling; commercial aircraft cabin redesign; spacecraft mission hardware rollout; rocket engine test launch \\
 & Consumer Goods & food beverage packaging redesign; cosmetics packaging redesign launch; household brand packaging redesign; brand logo identity unveiling \\
\midrule
\multirow{2}{*}{Culture}
 & Art & gallery exhibition artwork debut; public art installation unveiling \\
 & Architecture & cultural building design unveiling; new stadium design plans \\
\bottomrule
\end{tabular}
\end{table}

\section{Evaluation Details and Judge Prompts}
\label{sec:app_metrics}

\paragraph{Search score.}
The VLM compares the agent's references with the ground-truth image and
returns a binary verdict for each query. A pass requires a usable view of the
correct target, with identifying details clear enough to guide generation.
Differences in viewpoint or background are allowed; a wrong version or an
unusable view fails. Failed rollouts and missing references count as failures;
judge errors are excluded from the denominator.

\paragraph{Generation score.}
The per-prompt score is
\begin{equation}
    100\,(0.8\,s_{\mathrm{id}} + 0.2\,s_{\mathrm{ctx}}).
\end{equation}
The identity score $s_{\mathrm{id}}$ checks whether $e$ is drawn correctly: the
VLM compares $I_{\mathrm{gen}}$ with the ground-truth reference and assigns 1,
0.2, or 0 for consistent, partially consistent, or inconsistent identity. When a
prompt contains multiple targets, each target is scored separately and
$s_{\mathrm{id}}$ is their average. The context score $s_{\mathrm{ctx}}$ is the
fraction of rubrics, covering the scene, relations, and other context in $p$,
that the VLM judges as satisfied. The prompt below defines the identity tiers
and context judgments.

\clearpage
\subsection{Search Judge Prompt}
The judge receives the original user prompt, followed by the ground-truth
image and the agent's reference images, with the following system prompt.
\begingroup
\small
\setlength{\tabcolsep}{10pt}
\setlength{\LTpre}{6pt}
\setlength{\LTpost}{8pt}
\setlength{\parindent}{0pt}
\begin{longtable}{@{}p{\linewidth}@{}}
\caption{System prompt for the search judge.}\label{tab:search_judge_prompt}\\
\toprule
\rowcolor{black!6}[0pt][0pt]\textbf{Search judge} \hfill \textsf{System message}\\
\midrule
\endfirsthead
\multicolumn{1}{@{}l@{}}{\small\itshape Table \thetable{} (continued)}\\
\toprule
\rowcolor{black!6}[0pt][0pt]\textbf{Search judge} \hfill \textsf{System message}\\
\midrule
\endhead
\midrule
\multicolumn{1}{@{}r@{}}{\footnotesize\itshape Continued on next page}\\
\endfoot
\bottomrule
\endlastfoot
You are grading a retrieval step inside an image-generation pipeline. An agent was given a prompt, searched the web, and returned a set of REFERENCE IMAGES that will be handed to an image-generation model together with that prompt.
\tabularnewline\addlinespace[5pt]
You also get the GROUND-TRUTH image: the picture that provably shows the correct appearance of the target named by the prompt.
\tabularnewline\addlinespace[5pt]
ONE QUESTION: if the image model receives the agent's images instead of the ground-truth image, will it draw the target correctly?
\tabularnewline\addlinespace[5pt]
\textbf{JUDGE FROM THE POINT OF VIEW OF DRAWING, NOT PIXEL SIMILARITY.}\par
\begin{itemize}
\setlength{\itemsep}{2pt}\setlength{\parsep}{0pt}
\item PASS is correct when the agent's images show the SAME thing as the ground truth, from any angle, distance, crop, lighting or background, and clearly enough that its identifying details could be reproduced. A press photo, a product shot, an official render and a leaked picture of the very same design are interchangeable here.
\item PASS is still correct when the agent's images add extra content, or show the target as part of a larger scene, as long as the target itself is visible and legible.
\item FAIL when the target in the agent's images is a DIFFERENT thing: another edition, year, model, trim, colourway, person or named item. Near-identical siblings from the same series are the most important failures to catch --- they look right and steer the generation wrong.
\item FAIL when the agent returned no usable view of the target: only logos, crowds, venues, packaging, web-page screenshots, unrelated stock imagery, or a view too small, dark or obstructed for its details to be read.
\item FAIL when the target is absent even though the images are otherwise about the right event.
\end{itemize}
\tabularnewline\addlinespace[5pt]
Extra images that are irrelevant do NOT cause a failure by themselves, as long as at least one image lets the target be drawn correctly. But if the set as a whole would most likely lead the model to draw the wrong version, that is a FAIL.
\tabularnewline\addlinespace[5pt]
IMAGE ORDER: image 0 is the GROUND TRUTH. The remaining images, in order, are the agent's reference images.
\tabularnewline\addlinespace[5pt]
\textbf{Return JSON only:}\par
\begin{minipage}{\linewidth}
\begin{lstlisting}[style=judgejson]
{"target": "<2-8 words: the thing the prompt actually asks to be drawn>",
 "gt_shows": "<2-8 words: what the ground-truth image shows of that target>",
 "agent_shows": "<2-8 words: what the agent's images show of that target>",
 "pass": true|false,
 "why": "<one short sentence: the concrete evidence -- the named difference if you fail it>"}
\end{lstlisting}
\end{minipage}
\tabularnewline\addlinespace[5pt]
\end{longtable}
\endgroup

\clearpage
\subsection{Generation Judge Prompt}
The judge receives the target description, original user prompt, context
rubrics, generated image, and ground-truth images. Each image is explicitly
marked as generated or ground truth. The system prompt is reproduced below.
\begingroup
\small
\setlength{\tabcolsep}{10pt}
\setlength{\LTpre}{6pt}
\setlength{\LTpost}{8pt}
\setlength{\parindent}{0pt}
\begin{longtable}{@{}p{\linewidth}@{}}
\caption{System prompt for the generation judge.}\label{tab:generation_judge_prompt}\\
\toprule
\rowcolor{black!6}[0pt][0pt]\textbf{Generation judge} \hfill \textsf{System message}\\
\midrule
\endfirsthead
\multicolumn{1}{@{}l@{}}{\small\itshape Table \thetable{} (continued)}\\
\toprule
\rowcolor{black!6}[0pt][0pt]\textbf{Generation judge} \hfill \textsf{System message}\\
\midrule
\endhead
\midrule
\multicolumn{1}{@{}r@{}}{\footnotesize\itshape Continued on next page}\\
\endfoot
\bottomrule
\endlastfoot
You are an objective AI image evaluator. For one GENERATED IMAGE you produce two things: a TARGET CONSISTENCY TIER, and true/false verdicts on a list of CONTEXT criteria.
\tabularnewline\addlinespace[5pt]
\textbf{You receive:}\par
\begin{enumerate}
\setlength{\itemsep}{2pt}\setlength{\parsep}{0pt}
\item The generated image. This is the ONLY image being scored.
\item One or more ground-truth photographs of the real target. They show what the target actually looks like. They are never the image being scored.
\item The name of the target and the original prompt.
\item CONTEXT criteria about the requested action, placement, setting, surrounding objects, counts, or spatial relationships.
\end{enumerate}
\tabularnewline\addlinespace[5pt]
\textbf{Part 1: target consistency tier}
\tabularnewline*\addlinespace[5pt]
The question is exactly this: is the target shown in the generated image THE SAME THING as the target shown in the ground-truth photographs?
\tabularnewline\addlinespace[5pt]
It is NOT "do the two images look alike". The prompt deliberately asks for a different scene from the ground-truth photographs: a different moment, pose, camera angle, background, lighting and framing. That is by design. Those differences NEVER affect the tier.
\tabularnewline\addlinespace[5pt]
Judge only identity-bearing features: text and its spelling, logo and emblem shape and where it sits on the target, colour scheme and how the colours are distributed, model or variant silhouette, patterns and textures, materials, distinctive structural parts.
\tabularnewline\addlinespace[5pt]
\textbf{Procedure:}\par
\renewcommand{\labelenumi}{\alph{enumi}.}
\begin{enumerate}
\setlength{\itemsep}{2pt}\setlength{\parsep}{0pt}
\item Read the ground-truth photographs and list the target's identity-bearing features. List only features actually visible in the ground-truth photographs.
\item For each listed feature, check the generated image and sort it into matched or mismatched.
\item Assign the tier.
\end{enumerate}
\tabularnewline\addlinespace[5pt]
\textbf{Tier 1.0, CONSISTENT.}\par
Every salient identity-bearing feature agrees. Someone familiar with this target would recognise it as this exact one. Rendering quality loss, stylisation and softened minor detail are acceptable: small text that is blurry but in the right place and the right colour still counts as matched. The question is "is this the same entity", not "is this pixel-identical".
\tabularnewline\addlinespace[5pt]
\textbf{Tier 0.2, PARTIALLY CONSISTENT.}\par
The generated image is clearly attempting this target and gets at least one SPECIFIC identity feature right (the main colour scheme, the overall silhouette, the position of one marking), but at least one hard error makes it impossible to accept as the same entity. Hard errors: text misspelled or replaced by a different word, a logo replaced by a different logo or with a clearly different shape, the dominant colour scheme wrong, a clearly different model or variant.
\tabularnewline\addlinespace[5pt]
\textbf{Tier 0.0, INCONSISTENT.}\par
The target is absent, or a different entity is depicted, or nothing beyond generic category membership agrees.
\tabularnewline\addlinespace[5pt]
\textbf{Hard rules for the tier. Apply them literally:}\par
\begin{itemize}
\setlength{\itemsep}{2pt}\setlength{\parsep}{0pt}
\item Matching only the generic category is 0.0, NOT 0.2. "It is a cycling jersey", "it is a race car", "it is a necklace" earns nothing on its own. Tier 0.2 requires at least one specific identity feature to agree.
\item Scene, background, pose, camera angle, lighting, framing and crop differences never change the tier.
\item A feature not visible in the ground-truth photographs is outside the judgement. Do not list it and do not penalise it.
\item If the generated image crops the target but every visible part agrees, judge the visible part; tier 1.0 stays available.
\item If the target appears more than once in the generated image, judge the clearest instance.
\end{itemize}
\tabularnewline\addlinespace[5pt]
Emit one entry per distinct target entity named by the target description. When only one entity is named, emit exactly one entry.
\tabularnewline\addlinespace[5pt]
\textbf{Part 2: context criteria}
\tabularnewline*\addlinespace[5pt]
Judge each context criterion solely from the generated image. Ignore all content in the ground-truth photographs, including their background. Enforce the requested scene, action, placement, counts, other objects and spatial relations. Missing, wrong, illegible or unverifiable is false.
\tabularnewline\addlinespace[5pt]
\textbf{Output}
\tabularnewline*\addlinespace[5pt]
Return a JSON object with exactly these keys:
\par
\begin{minipage}{\linewidth}
\begin{lstlisting}[style=judgejson]
{
  "entities": [{
    "name": "...",
    "gt_features": ["..."],
    "matched": ["..."],
    "mismatched": ["..."],
    "tier": 1.0,
    "reason": "at most 20 words"
  }],
  "context_results": [true, false],
  "context_reasons": ["short evidence", "short evidence"]
}
\end{lstlisting}
\end{minipage}
\tabularnewline\addlinespace[5pt]
\texttt{tier} must be exactly 1.0, 0.2 or 0.0. \texttt{context\_results} and \texttt{context\_reasons} must each have exactly as many entries as the context criteria list, in the same order.
\tabularnewline\addlinespace[5pt]
\end{longtable}
\endgroup

\subsection{Introduction to Mind-Bench and KnowGen}
\label{sec:app_transfer_metrics}
\paragraph{Mind-Bench.}
Mind-Bench~\citep{he2026mindbrush} contains 500 prompts across ten equally sized categories, divided into Knowledge and Reasoning. A VLM checks each generated image against a sample-specific checklist; a sample passes only if every item is satisfied. We report pass rates for both groups and overall.

\paragraph{KnowGen.}
KnowGen~\citep{feng2026gensearcher} contains 630 prompts in Science \& Knowledge and Pop Culture \& News. A VLM evaluates visual correctness, text accuracy, prompt faithfulness, and aesthetics on a $\{0,0.5,1\}$ scale. K-Score weights these dimensions by 0.4, 0.4, 0.1, and 0.1, respectively, and averages the two groups equally. We report all scores on a 0--100 scale.

\section{Prospective Validation of the Three-Day Retrieval Cutoff}
\label{sec:app_time_window}

In \cref{sec:query_synthesis}, we treat the news date as the issue time $\tau$ and restrict subsequent text and image searches to results published no later than $\tau+3$ days. This date filtering approximates the web environment shortly after an event, when newly introduced entities exhibit temporal novelty. We assess this approximation by comparing image-search results collected in real time with those retrieved later under a three-day date cutoff.

We prospectively tracked 50 entities extracted using our pipeline to measure changes in visual-reference retrieval during the first week after release. For each entity, we verified its release time $\tau$ from its initial public announcement and fixed an image-search query at enrollment. We collected the top eight results within 24 hours of release and again on days 3 and 7, without date filters. Each target was queried twice in each condition. We report Hit@8, the percentage of queries with at least one correct visual reference among the top eight results, averaged over the two repetitions.

To assess retrospective date filtering, we also queried the same targets on day 7 with the same fixed queries, restricting results to those published no later than $\tau+3$ days. We queried each target twice in this condition and compared the results with those collected on day 3 without date filters.

\begin{table}[H]
    \centering
    \small
    \setlength{\tabcolsep}{6pt}
    \renewcommand{\arraystretch}{1.15}
    \caption{Visual-reference retrieval for 50 prospectively tracked entities. Conditions A--C use searches without date filters at the indicated times; condition D searches on day 7 for results published no later than day 3. Release time $\tau$ is verified from each entity's initial public announcement.}
    \label{tab:retrieval_time_window}
    \begin{tabular}{@{}lllr@{}}
        \toprule
        Condition & Actual retrieval time & Publication-date cutoff & Hit@8 ($\%$) $\uparrow$ \\
        \midrule
        A: Initial release & Within 24 hours of $\tau$ & None & 68.0 \\
        B: Day 3 & $\tau + 3$  & None & 76.0 \\
        C: Day 7 & $\tau + 7$  & None & 80.0 \\
        \midrule
        D: Retrospective cutoff & $\tau + 7$  & $\tau + 3$  & 78.0 \\
        \bottomrule
    \end{tabular}
\end{table}

As shown in \cref{tab:retrieval_time_window}, Hit@8 increased from 68.0\% within 24 hours of release to 76.0\% on day 3 and 80.0\% on day 7. Correct visual references thus became easier to retrieve over the first week in this sample. Searching on day 7 with the three-day publication-date cutoff yielded 78.0\%, close to the 76.0\% observed on day 3. This comparison supports the use of date filtering to approximate visual-reference retrieval shortly after an event, consistent with the temporal-novelty setting used in our synthesis pipeline.

\section{Human-Model Alignment}
\label{sec:app_human_agreement}

We validate the VLM evaluator against human judgments using 100 images selected from \ourbench{}. Human annotators assess the search score and the identity and context components of the generation score, following the criteria in \cref{sec:app_metrics}. Search judgments check whether retrieved references match the ground-truth target; generation judgments assess target identity in the generated image and satisfaction of the context rubrics.

\begin{table}[H]
    \centering
    \small
    \setlength{\tabcolsep}{8pt}
    \renewcommand{\arraystretch}{1.15}
    \caption{Agreement between human annotators and the VLM evaluator in the 100-image validation study on \ourbench{}. Identity and context are the two components of the generation score.}
    \label{tab:human_vlm_agreement}
    \begin{tabular}{@{}llr@{}}
        \toprule
        Metric & Judgment & Agreement (\%) \\
        \midrule
        Search score & Reference matches ground truth & 98.0 \\
        Generation score: identity & Target identity matches ground truth & 98.0 \\
        Generation score: context & Context rubric satisfaction & 94.5 \\
        \bottomrule
    \end{tabular}
\end{table}

As shown in \cref{tab:human_vlm_agreement}, the VLM agrees with human annotators on 98\% of search judgments and 98\% of generation identity judgments. Agreement on context rubrics is slightly lower at 94.5\%, but remains high despite the finer-grained criteria. These results support the use of a strong teacher VLM to judge retrieval correctness against ground-truth references and to evaluate target identity and contextual details on \ourbench{}.

\clearpage
\section{Comparison of Agent Trajectories}
\label{sec:app_trajectories}

\begin{figure}[H]
    \centering
    \includegraphics[width=\linewidth,height=0.79\textheight,keepaspectratio]{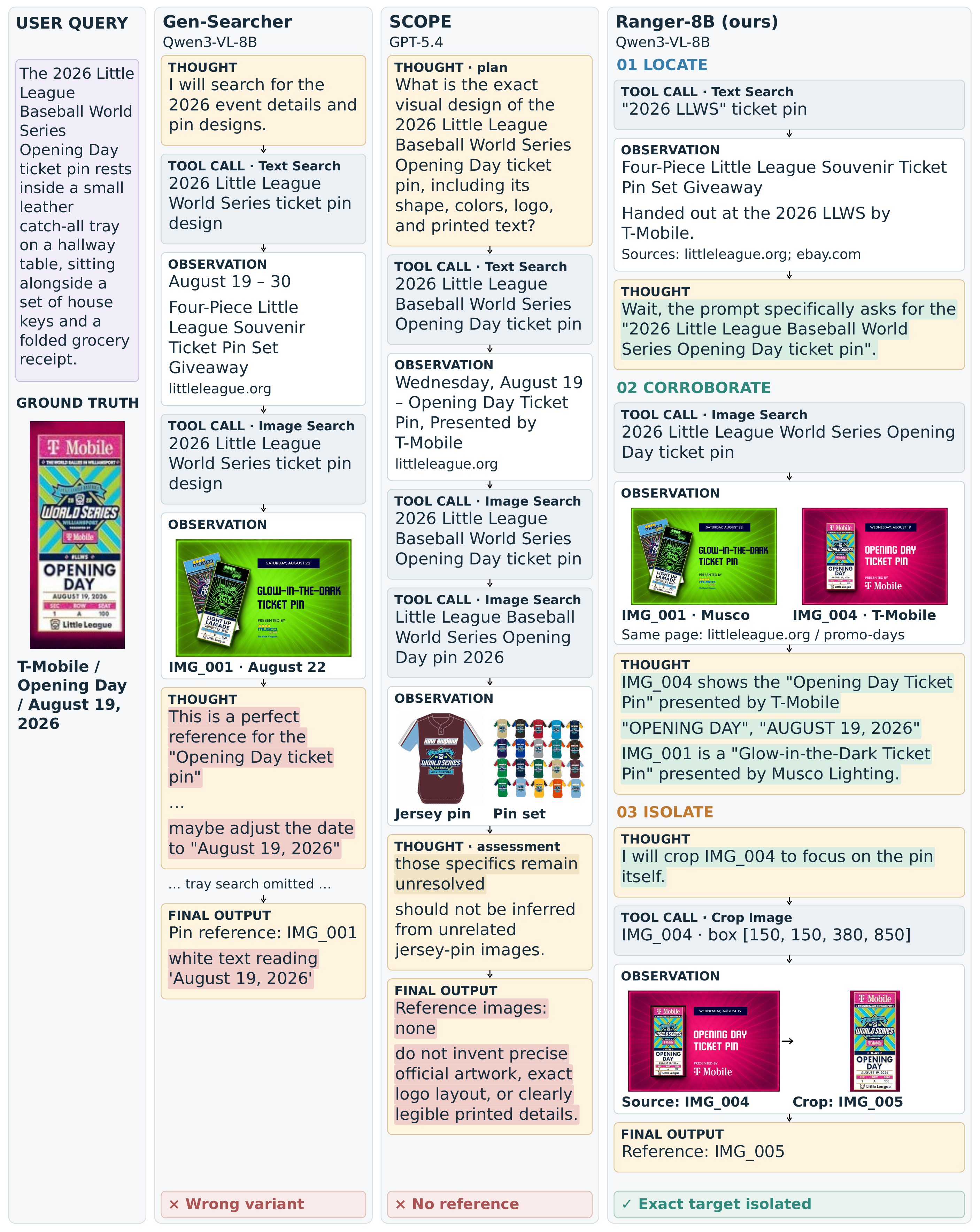}
    \caption{Pre-generation trajectories for the same target. Excerpts are verbatim; tool names are normalized and verdicts annotated. Green/amber/red mark grounded decisions, uncertainty, and incorrect or missing references. These independent runs omit SCOPE's later verification and repair. Ground truth is shown for comparison only.}
    \label{fig:app_trajectory_comparison}
\end{figure}

\clearpage
\section{Qualitative Generation Comparison on \ourbench{}}
\label{sec:app_cases_main}

\begin{figure}[H]
    \centering
    \includegraphics[width=\linewidth,height=0.79\textheight,keepaspectratio]{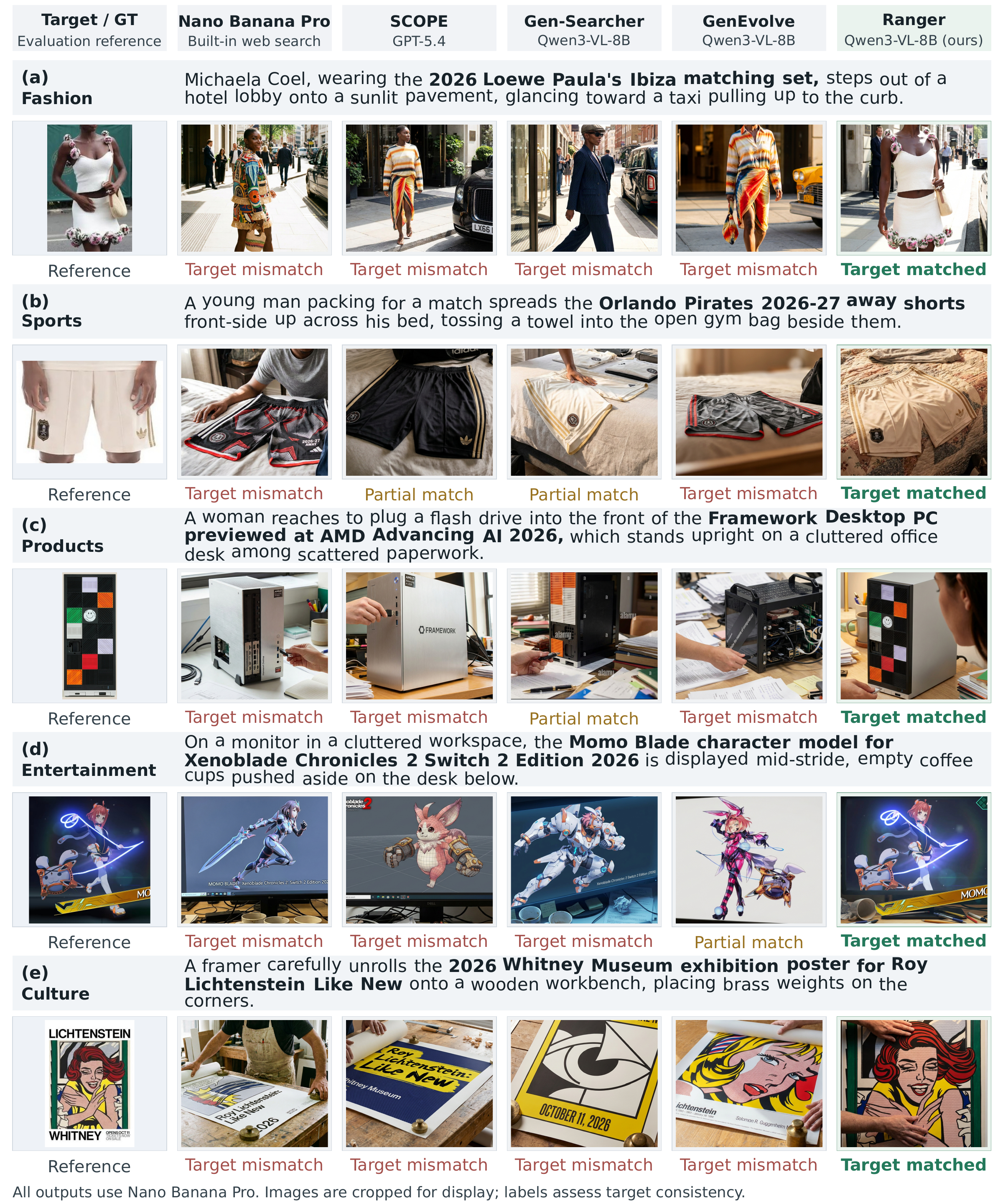}
    \caption{Generation comparisons across all five \ourbench{} domains, using Nano Banana Pro with built-in search or agent-selected references (SCOPE: GPT-5.4). Ground truth is for evaluation only. Labels assess target identity on original outputs; display crops emphasize the target.}
    \label{fig:app-generation-comparison-1}
    \label{fig:app-generation-comparison}
\end{figure}

\clearpage
\section{Gallery of Generated Images}
\label{sec:app_generation_gallery}

\begin{figure}[H]
    \centering
    \includegraphics[width=\linewidth,height=0.86\textheight,keepaspectratio]{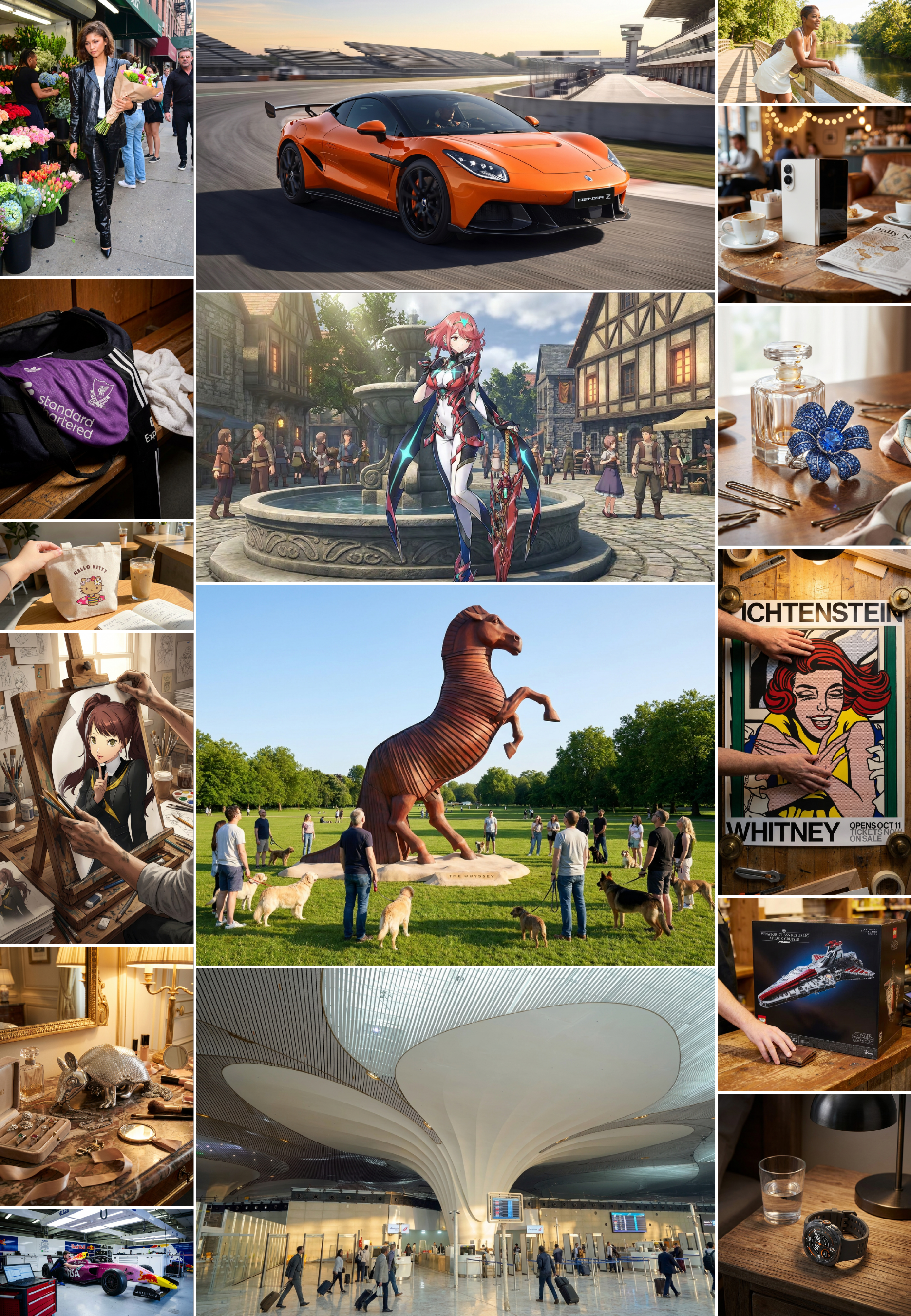}
    \caption{Sixteen Nano Banana Pro outputs from \ourmodel{} prompts and references, spanning all five \ourbench{} domains. Images are shown uncropped at their original aspect ratios.}
    \label{fig:app_generation_gallery}
\end{figure}
\clearpage
\section{Effect of Reference Cropping}
\label{sec:app_crop_cases}

\begin{figure}[H]
    \centering
    \includegraphics[width=\linewidth,height=0.75\textheight,keepaspectratio]{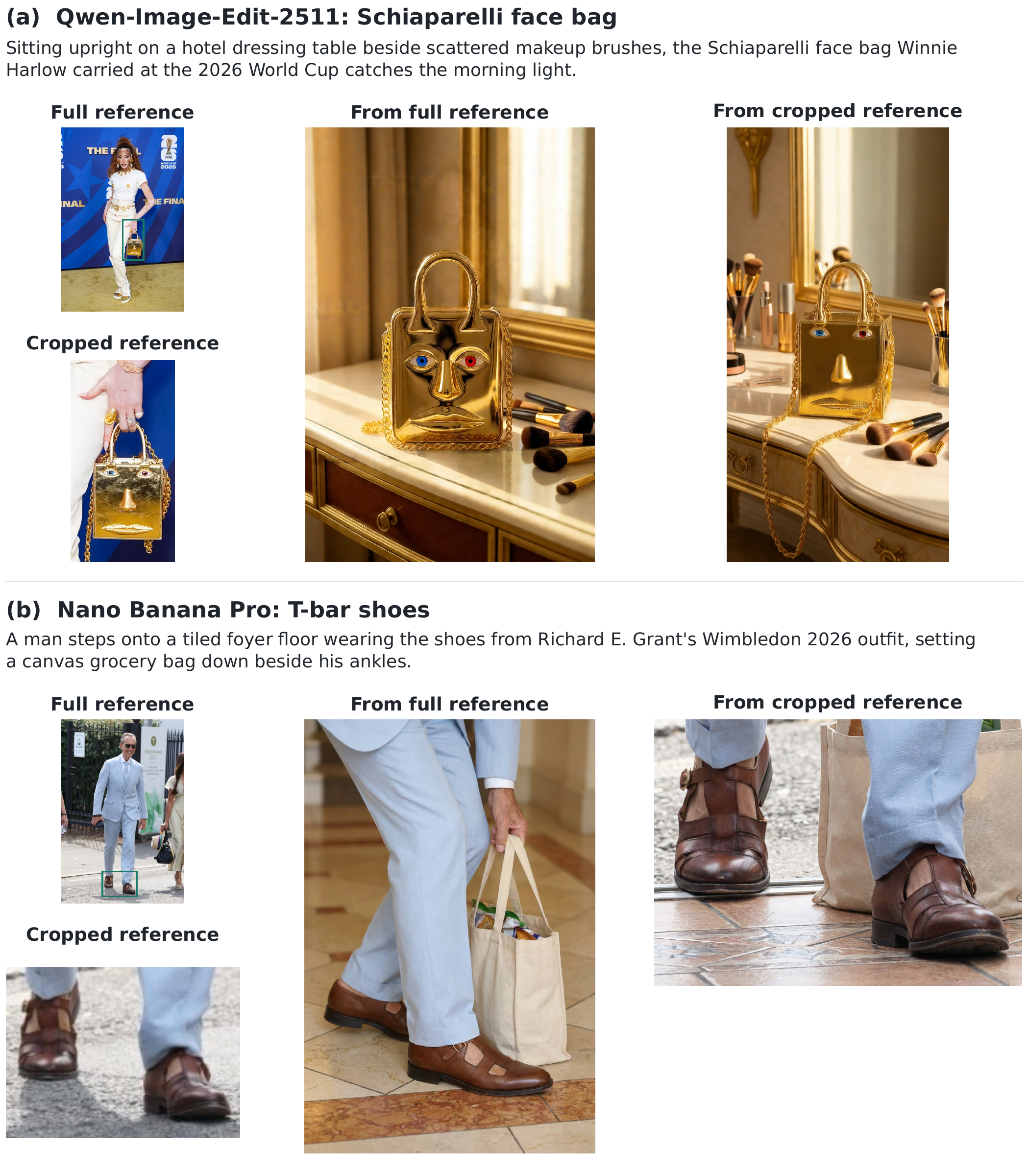}
    \caption{Reference-cropping comparisons with Qwen-Image-Edit-2511 and Nano Banana Pro. Each case shows the user prompt, full and cropped references, and paired generations. Green boxes mark the reference crops.}
    \label{fig:app_crop_comparison}
\end{figure}

With the full reference, Qwen-Image-Edit adds prominent eyebrows and oversized facial features to the bag; the crop yields a flatter face with smaller eyes. For Nano Banana Pro, the crop restores the shoes' central T-bar and horizontal straps, replacing the isolated side cutouts in the full-reference output.

\end{document}